\documentclass[11pt]{article}

\usepackage[final]{acl}

\usepackage{times}
\usepackage{latexsym}
\usepackage{booktabs}
\usepackage{amsmath}
\usepackage{listings}
\usepackage[table]{xcolor}
\usepackage{pifont}
\usepackage{tabularx}
\usepackage{ragged2e} 
\usepackage{enumitem}

\usepackage{listings}

\usepackage{placeins}

\usepackage[T1]{fontenc}

\usepackage[utf8]{inputenc}

\usepackage{microtype}

\usepackage{inconsolata}

\usepackage{graphicx}

\author{%
  Siyuan Zheng\thanks{These authors contributed equally.} \quad
  Yifan Duan\footnotemark[1] \quad
  Chao Xue \quad
  Flora D.\ Salim\thanks{Corresponding author.}\\[2pt]
  UNSW Sydney\\[2pt]
  \texttt{siyuan.zheng1@student.unsw.edu.au}\\
  \texttt{xuechao8071@gmail.com}\\
  \texttt{\{yifan.duan3,\,flora.salim\}@unsw.edu.au}
}

\title{Scope3Trace: Evidence-Based Identification and Extraction of Scope 3 GHG Emissions from Sustainability Reports
}

\begin{document}
\maketitle

\begin{abstract}

Scope~3 greenhouse gas (GHG) emissions account for the majority of corporate carbon footprints, yet remain difficult to analyze at scale due to sparse disclosures, heterogeneous report document formats, and limited evidence traceability. Existing approaches typically rely on large language models to extract emissions information from ESG reports, but often lack explicit evidence grounding or depend on costly manual annotation and verification to ensure extraction reliability. To address these challenges, we propose \textbf{Scope3Trace}, an evidence-grounded information extraction framework designed to extract interpretable and traceable Scope~3 emissions information from real-world ESG and sustainability reports. The framework integrates a document information extraction pipeline that performs PDF collection and OCR parsing, LLM-assisted page localization and table reconstruction, and hybrid rule–LLM extraction of organization- and building-level emissions disclosures with evidence-grounded verification. Building upon this framework, we further contribute a dual-level, evidence-grounded, multimodal dataset comprising organization-level Scope 3 disclosures extracted from heterogeneous sustainability reports. \textbf{Scope3Trace} enables reliable extraction and transparent integration of heterogeneous sustainability disclosures, achieving high accuracy in extracting Scope~1–3 totals and category-level disclosures from sustainability reports.
The code is available at \url{https://github.com/cruiseresearchgroup/Scope3Trace}.
\end{abstract}

\section{Introduction}

\begin{figure}[t]
    \centering
    \includegraphics[width=0.95\linewidth]{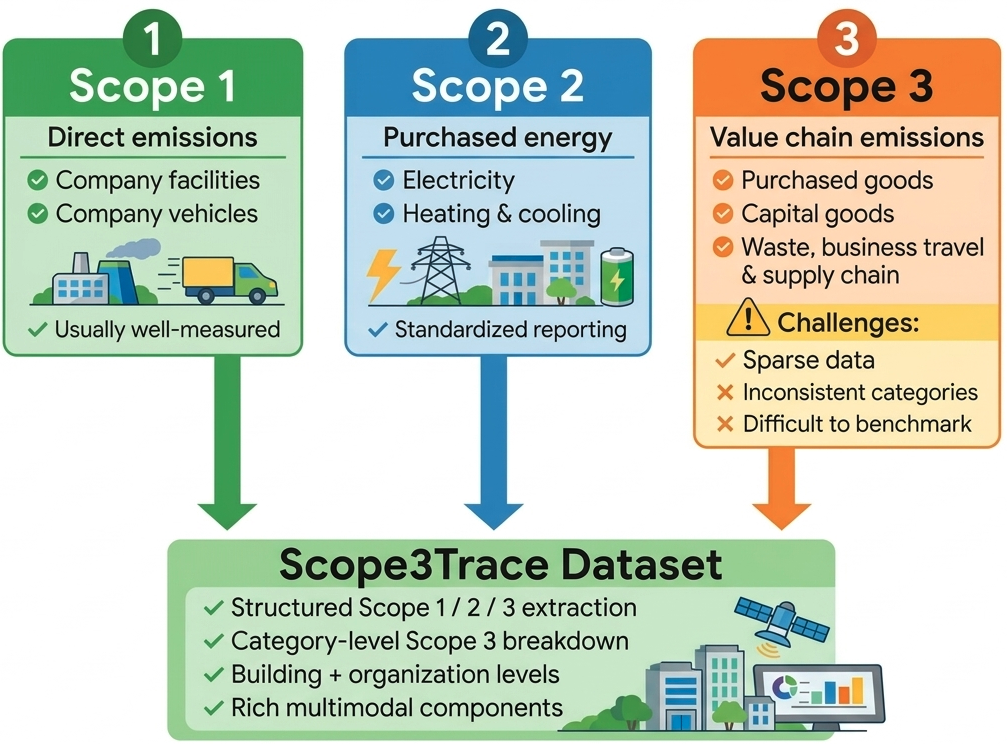}
    \caption{Illustration of Scope 1,2,3 emissions under the GHG Protocol. Using our information extraction framework, we construct the Scope3Trace dataset that organizes extracted emissions disclosures across organization and building levels with evidence.
    }
    \label{fig:intro_scope3trace}
\end{figure}

As climate change increasingly threatens global sustainability and economic stability, corporate greenhouse gas (GHG) reporting has emerged as a cornerstone of climate governance, enterprise risk management, and regulatory accountability \citep{traub2025carbon,luu2025mandatory,li2025carbon}. Within corporate carbon accounting frameworks, emissions are categorized into Scope 1, Scope 2, and Scope 3 based on their source and value-chain boundary, as seen in Figure~\ref{fig:intro_scope3trace}. While existing research and industry efforts have increasingly focused on Scope 1 and Scope 2 emissions due to their standardized reporting protocols and relatively accessible activity data \citep{Yap2025BuildingCarbon}, Scope 3 emissions, despite accounting for the majority of corporate carbon footprints, remain difficult to analyze in data-driven studies due to fragmented disclosures and the lack of unified structured datasets derived from sustainability reports \citep{ghgprotocol2019scope3,Hertwich2018Scope3,GHGProtocol2011}.

Recent advances in large language models (LLMs) have significantly improved document understanding capabilities \citep{achiam2023gpt}. In addition, retrieval-augmented generation (RAG) techniques help mitigate hallucination by grounding generation in retrieved external evidence \citep{gao2023retrieval}. Recent studies have begun to leverage these capabilities to extract corporate emissions disclosures from sustainability and ESG reports using LLM-based information extraction pipelines, often combined with manual verification or annotation to ensure extraction reliability \citep{Beck2025}. In parallel, another line of work leverages LLMs to infer a company’s industry sector from corporate reports, and subsequently estimates company-level emissions by combining sector-level emission factors with company activity or spending data \citep{Dumit2024Atlas,Guo2025ExioNAICS,Lenzen2012MRIO}.

While promising, these approaches leave important challenges unresolved. Extraction-focused methods \citep{Beck2025} improve access to disclosed figures but often overlook reporting boundaries, asset coverage, and evidence traceability. Besides, these works frequently rely on costly and time-consuming manual expert verification to ensure extraction reliability. Estimation-based approaches \citep{Guo2025ExioNAICS} provide broader coverage but abstract away from organization- or asset-level characteristics, offering limited transparency and weak alignment with disclosed accounting practices. As a result, Scope 3 data remains difficult to interpret, validate, and operationalize for decision-making.

To address these limitations, we present \textbf{Scope3Trace}, an evidence-grounded information extraction framework for sustainability and ESG reports. The framework implements a document extraction pipeline that collects sustainability reports, converts PDFs into structured OCR text, and identifies pages containing Scope~1–3 disclosures through page-level localization. It then reconstructs tabular disclosures using LLM-based table interpretation and performs structured extraction of emissions values through a hybrid rule–LLM design, where LLMs assist in page localization and table reconstruction, while rule-based components handle numeric parsing, category mapping, and evidence verification against the OCR text. This design enables reliable extraction of organization- and building-level emissions disclosures with explicit textual evidence grounding. Using this framework, we further organize the extracted records into a dual-level dataset spanning organization- and building-level Scope emissions information.

Our key contributions can be summarized as follows:

\begin{itemize}

\item 
We propose an evidence-grounded information extraction framework for sustainability reports that integrates PDF collection, OCR-based document parsing, LLM-assisted page localization and table reconstruction, and hybrid rule--LLM structured extraction for Scope~1--3 emissions disclosures.

\item 
We design a reliable extraction pipeline that explicitly links extracted emissions values to supporting textual evidence, enabling traceable and verifiable extraction of organization- and building-level Scope emissions information from heterogeneous ESG reports. 

\item 
To the best of our knowledge, \textit{Scope3Trace} is among the first efforts to construct a dual-level, multimodal, and evidence-grounded dataset specifically targeting Scope 3 emissions.

\end{itemize}

\section{Related Work}

\textbf{Information Extraction from ESG and Sustainability Reports.} Recent work has shown that NLP and LLM-based systems can reliably extract structured emissions information from sustainability and ESG reports, including Scope~1--3 values, standardized categories, and supporting textual or tabular evidence \citep{Beck2025,Mishra2024Statements,Ding2025EulerESG}. These approaches substantially reduce manual reporting effort and form the basis of large-scale access to disclosed emissions data. Their task definitions and evaluations, however, are primarily document-centric, focusing on extraction accuracy and evidence localization rather than how reported values relate to underlying assets or organizational coverage.

\textbf{Scope 3 Emission Estimation and Proxy-Based Methods.} A parallel line of research estimates Scope~3 emissions using economic, activity-based, or spend-based proxies, often relying on input--output models or sectoral emission factors to achieve broad organizational coverage \citep{Hertwich2018Scope3,Dumit2024Atlas,Guo2025ExioNAICS,duan2026ghgbench}. Such methods are well suited for large-scale benchmarking and policy analysis, but typically abstract away from asset-level structure and provide limited traceability to disclosed figures or reporting boundaries.

\textbf{Building and Asset-Level Carbon Modeling.} Carbon modeling at the level of individual buildings and physical assets has enabled detailed analyses of operational and embodied emissions using building attributes, energy data, and geospatial context \citep{Yap2025BuildingCarbon}. While these studies offer strong physical interpretability and urban-scale insights, they are generally developed independently of corporate sustainability disclosures and Scope~3 reporting frameworks.

\textbf{Data Quality, Coverage, and Practical Usability in ESG Systems.} Prior work has highlighted persistent challenges in ESG data quality, including incomplete disclosures, inconsistent boundaries, and heterogeneous estimation practices, particularly for Scope~3 emissions \citep{Beck2025,dimmelmeier-etal-2024-informing,sternberg2022cop27}. Although these limitations are widely acknowledged, they are rarely encoded explicitly in machine-readable form, leaving coverage, applicability, and confidence largely implicit in existing ESG datasets and systems.

\textbf{Reliability, Hallucination, and Trustworthiness of LLMs.} Since our framework relies on LLMs for extraction and verification, the reliability of LLM-based systems is directly relevant to our design. Recent studies have examined when LLMs fabricate citations under deployment constraints \citep{zhao2026deployment}, proposed white-box hallucination detection based on per-layer representations \citep{yang2026trilens}, and provided a comprehensive survey of trustworthiness in LLM reasoning \citep{wang2025comprehensive}. These works motivate our evidence-grounded protocol, which pairs every extracted value with verifiable textual evidence rather than relying solely on model outputs.

\textbf{Retrieval-Augmented Generation and Multi-Agent Systems.} Beyond single-model extraction, RAG and multi-agent frameworks have become important paradigms for grounding and coordinating LLM behavior. Recent work studies transparent handling of knowledge conflicts in RAG \citep{ye2026seeing}, measures the utility of retrieved evidence through semantic perplexity reduction \citep{dai2025seper}, and analyzes LLM-oriented retrieval from a denoising-first perspective \citep{dai2026llm}. Complementary studies combine retrieval with multi-agent multimodal reasoning for verification \citep{li2026retrieval} and stabilize multi-passage retrieval through fusion controllers \citep{zhu2026fcpragfusioncontrollerparametricretrievalaugmented}. Reinforcement learning has likewise been applied to strengthen multi-hop reasoning over structured knowledge in question answering \citep{wen2026reinforcement}. Multi-agent designs further improve reliability through role decomposition, chain-of-draft reasoning, and associative memory \citep{li2026draftrl,zhu2026hela,li-etal-2026-zara}. These advances inform our multi-agent pipeline, in which specialized agents share intermediate state and cross-validate one another's outputs.

\textbf{Broader Applications.} The LLM and deep-learning techniques underpinning our framework have been validated across a broad range of LLM-related applications \citep{kuang2025learning,nie2026understanding,li2026boostapr,tan2026evocurl,zhu2026learninglookagainlossgap,zhou2026comem,zhu2026doeslongcontextsupervisionactually,11460600,li-etal-2025-sensorllm}. Their wide adoption across such diverse tasks underscores the reliability and generalizability of these methods, which have become standard building blocks of modern text and multimodal processing \citep{yu2025test,10.1007/978-981-95-4384-7_10,du2026possibledefinitebenchmarkevaluating,yuan2026sp,10.1007/978-981-95-4100-3_2,qi2025over++}. Beyond language-centric settings, the underlying machine-learning techniques, covering representation learning, predictive modeling, and statistical estimation, have also proven effective in a broad range of other application domains \citep{xiao2025multifrequency,lu2026s,hongmo2026lightweight,shao2026continuous,shao2026design}. This breadth reflects the maturity of the text processing and learning techniques that our extraction framework builds upon \citep{sheng2026group,8615588,zhang2025lightweight,xiao2026points}.


\begin{figure*}[t]
    \centering
    \includegraphics[width=0.925\textwidth]{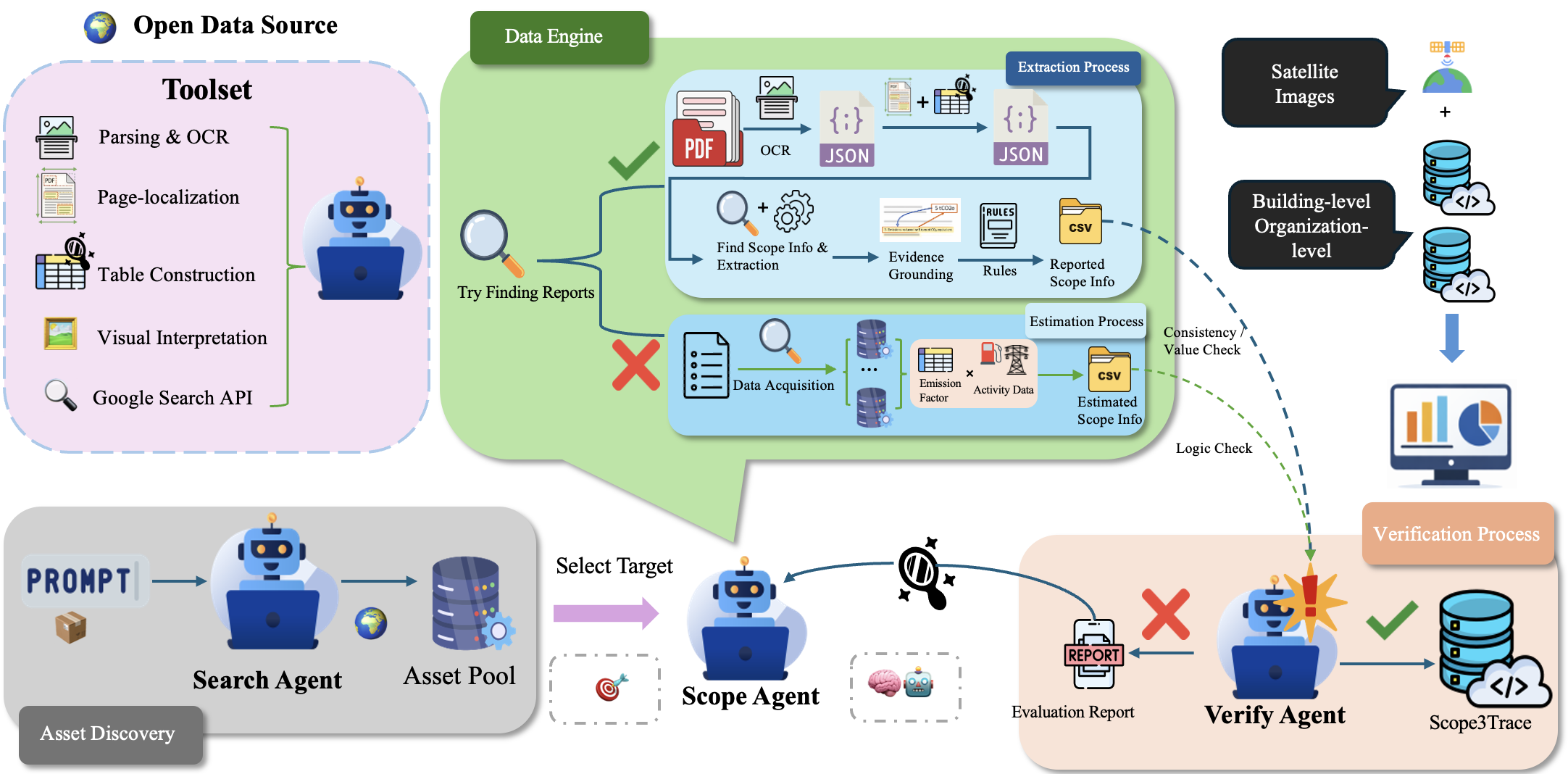}
    \caption{
Overview of the Scope3Trace multi-agent system for dataset construction. 
The pipeline is composed of three specialized agents: the Search Agent, Scope Agent, and Verify Agent.
to perform data acquisition, parsing, extraction, and verification. 
By combining structured parsing (e.g., page localization and table reconstruction) 
with explicit evidence grounding (sample in \ref{sam:smapleofbuildinglevel}), the system ensures that all extracted values 
are traceable and reliable, enabling audit-grade Scope~1/2/3 emissions data construction.
}
    \label{fig:multi-agent_overview}
\end{figure*}

\section{Scope3Trace}


This section presents the \textbf{Scope3Trace} framework: §\ref{sec:motivation} states the design goal, §\ref{sec:dataset} describes the dual-level data representation, and §\ref{sec:agentpipeline} introduces the multi-agent pipeline for data extraction and verification.



\subsection{Motivation and Design Goal}
\label{sec:motivation}

The fundamental limitations of Scope3 emissions estimation lie not only in numerical accuracy but also in the limited observability of activity data at the physical asset level, which is further exacerbated by the heterogeneity of unstructured ESG report texts—a core challenge that semantic and text processing technologies are uniquely positioned to address. In practice, organization-level disclosures are often embedded in unstructured paragraphs, tables, or footnotes of sustainability reports, with inconsistent wording, ambiguous category definitions, and varying document formats. including scanned PDFs, formatted text, and mixed text-image layouts. Existing studies either depend on manual annotation, which is expensive and difficult to scale, or adopt generic document processing pipelines that lack sufficient semantic understanding. As a result, the extracted emissions data suffer from weak traceability and interpretability.

To address these gaps, our design goal is to construct a dual-level Scope 3 dataset by leveraging advanced text processing and semantic techniques to automate the extraction, normalization, and verification of emissions information from heterogeneous ESG reports. Specifically, we aim to: (1) use semantic models to achieve accurate localization and interpretation of emissions-related content in unstructured texts/tables; (2) Develop rule-LLM hybrid pipelines to guarantee evidence grounding and traceability by connecting extracted values to their original text snippets; (3) apply semantic alignment to unify heterogeneous disclosures into a standardized schema, enabling interpretable, diagnosable, and coverage-aware emissions analysis. This design eliminates the over-reliance on manual annotation and enhances the scalability of the framework.

\subsection{Dataset Composition and Structure}
\label{sec:dataset}

Scope3Trace adopts a multi-level data structure explicitly optimized for information extraction and semantic alignment. This structure distinguishes enterprise-disclosed emissions data obtained from ESG reports through text processing techniques from asset-level representations. It is collaboratively generated by an enhanced multi-agent system and comprises two tightly coupled data layers: organizational-level disclosures retrieved from reports through information extraction, and building-level asset records augmented with semantic metadata produced by text analysis.

At the organizational level, each record corresponds to an organization and a specific reporting year. It contains extracted total Scope 3 emissions, category-level disaggregations with semantic labels normalized through text analysis, and standardized Scope 1 and Scope 2 emissions. The record also integrates metadata such as reporting boundaries and audit status, which are captured using entity recognition and relation extraction. A key characteristic of this layer is the evidence link: every extracted value is paired with a text segment from the original ESG report identified by semantic models, together with confidence scores produced by the extraction pipeline.

At the building level, each record corresponds to an individual building and year, documenting physical attributes, geospatial coordinates, and operational activity indicators. The records are further supplemented with semantic metadata, including building type classifications generated by text classification and activity description standards refined through semantic similarity computation. Multimodal satellite imagery is integrated as supplementary context, while the core descriptive metadata is enhanced via text processing to ensure alignment with organizational-level disclosures. Such alignment includes connecting building names mentioned in reports to corresponding asset entries through entity linking. All derived fields are clearly marked with their provenance, distinguishing information from direct extraction, modeling, or estimation, alongside confidence scores obtained from semantic processing. This labeling ensures transparency and prevents ambiguity with enterprise-reported data.

\subsection{Data Source and Construction}
\label{sec:agentpipeline}
Scope3Trace integrates organization-level disclosures and building-level inventories through a modular pipeline centered on information extraction and semantic processing as shown in Figure \ref{fig:multi-agent_overview}. The pipeline is composed of three specialized agents: the Search Agent, Scope Agent, and Verify Agent. As illustrated in Figure~\ref{fig:multi-agent_overview}, the workflow starts with data acquisition, proceeds through parsing, extraction, estimation, and verification, and finally outputs the integrated Scope3Trace results. To ensure transparency, reproducibility, and interpretability, the workflow combines rule-based text parsing with bounded LLM application that supports semantic understanding rather than speculative inference, aligning with the pipeline’s core steps, and each construction phase is assigned to a designated agent or collaborative effort. Restricting agent autonomy in this way is a deliberate, empirically validated design choice: under matched conditions (identical role decomposition, backbone LLM, and prompts), free-form conversational agent interaction repeatedly overwrites originally correct values (Appendix~\ref{app:autogen}). The agents therefore share intermediate state and cross-validate one another's outputs, but do not engage in open-ended conversational orchestration.
The construction proceeds in four phases, all underpinned by semantic and information extraction technologies and consistent with the pipeline’s process:
(1) Locating evidence by Search Agent: obtaining public reports via the
Google Search API, running PDF OCR parsing through the Parsing and OCR
Toolset, and performing emissions-related content localization (the
pipeline's Page-localization and Find Scope Info modules), returning a
verbatim evidence list without performing any extraction or computation;
(2) Extraction and estimation by Scope Agent: compiling the physical
asset inventory (the Asset Pool) with entity linking, extracting
structured values from the located evidence, and computing the
deterministic building-level estimates of Appendix~\ref{sec:formulas}
(the Estimation Process, integrating Activity Data and Emission
Factor); (3) Verification by Verify Agent: deterministic span-level
checking of every extracted value against the source OCR text, plus
semantic validation of rule applicability and boundary compatibility
(the Verification Process); (4) Cross-level alignment diagnostics by all
three agents: leveraging semantic similarity and relation extraction to
ensure consistency between levels, corresponding to the pipeline's
Consistency Value Check and Logic Check.

We experimented with three families of LLMs as candidates for the agents' underlying model, all under temperature=0. As shown by the single-call LLM rows of Table~\ref{tab:extraction_results}, the three models reach within 0.02 scope-level F1 of one another on the gold evaluation set, suggesting that the semantic portion of the extraction task is largely insensitive to LLM choice. This robustness reflects a deliberate design principle: deterministic logic such as numeric parsing, unit normalization, category mapping, and evidence checking is delegated to rule-based components, restricting the LLM to bounded semantic tasks. We adopt GPT-4o as the default backbone; swapping it across model families leaves performance within a narrow band (Appendix~\ref{app:backbone}), with per-agent cost in Appendix~\ref{app:overhead}.

\subsubsection{Building Asset Inventory}

Led by the Scope Agent with entity linking, text classification and semantic normalization modules, the building asset inventory integrates public building registries from Open Data Source, simulated energy activity datasets and disclosure-related asset information including Climate Active and building-tier reports, acting as the Asset Pool in the pipeline. Its core contribution is semantic alignment: the agent leverages pre-trained semantic models fine-tuned on ESG text corpora to link asset identifiers like building names and addresses in public registries with those appearing in the report evidence located by the Search Agent. It resolves ambiguities such as synonymous building names and abbreviations through contextual semantic understanding, laying the groundwork for later cross-level alignment.

Asset records are aligned via stable property identifiers, standardized addresses processed by address parsing and geospatial association methods. Deterministic parsers are used to supplement missing attributes, such as inferring building type from report text descriptions. The final inventory contains standardized locations, building features, multi-year activity indicators, optional satellite imagery, and semantic metadata including activity tags and asset category labels, supporting smooth alignment with organizational-level disclosures from ESG reports.

\subsubsection{Organization-level Inventory}

Parallel to the Scope Agent, the Search Agent handles document acquisition and evidence localization, executing the pipeline's Data Acquisition step and the localization stages of the Extraction Process. It first obtains public PDF sustainability and GHG inventory reports via the Google Search API. Documents then pass through a four-stage pipeline, of which the Search Agent executes stages 1--2 (parsing and localization) and the Scope Agent executes stages 3--4 (extraction and grounding), matching the pipeline's Parsing \& OCR Toolset, Page-localization, Table Construction, and Find Scope Info \& Extraction modules.

1. PDF-to-Text Parsing with Text Preprocessing: Following the Parsing \& OCR Toolset, it converts PDFs (including scanned files via OCR) into structured JSON raw text, then performs tokenization, stopword removal and sentence segmentation to clean the text for downstream tasks.

2. LLM-Based Emissions Content Localization: Directly corresponds to the pipeline’s Page-localization module, where we prompt an off-the-shelf LLM (GPT-4o by default, with temperature=0; see Appendix~\ref{app:backbone} for backbone robustness across model families) to perform page-level and sentence-level classification, identifying text snippets and tables relevant to Scope 1-3 emissions. This replaces generic page localization with semantic understanding, ensuring high recall of emissions-related content, which is a key prerequisite for subsequent extraction.

3. Semantic-Driven Table Reconstruction and Extraction: Aligns with the pipeline’s Table Construction and Find Scope Info \& Extraction steps, applying table structure recognition (TSR) combined with LLM-based table interpretation to extract numerical values, category labels, and contextual notes from tables. Rule-based parsers optimized for ESG report semantics handle numeric normalization including unit conversion and decimal alignment and category mapping such as unifying ``Scope 3 -- Purchased Goods and Services'' across different report wordings via semantic similarity, ensuring the extracted data is structured and consistent.

4. Evidence Grounding and Normalization: Corresponding to the pipeline’s Evidence Grounding Rules module, each extracted value is linked to its source text snippet identified via span detection and normalized into a unified schema using entity normalization. This ensures traceability and consistency across heterogeneous reports, which is a core requirement for the verification process. All pipeline rules, prompts, and thresholds were developed exclusively on a 20-PDF development set disjoint from the gold evaluation set in Section \ref{sec:extractionevaluation}.

The resulting organization-level disclosure dataset with one record per organization and reporting year includes Scope 1 and 2 totals, disclosed Scope 3 category breakdowns, reporting boundaries extracted via entity recognition, certification status, and evidence sources as text snippets, aligning with the pipeline’s Reported Scope Info output. The Scope Agent resolves inconsistencies such as mismatched asset names between report evidence and the asset inventory via semantic matching, ensuring the extracted data is consistent with its Asset Pool.

\subsubsection{Building-level Explanatory Components}

Scope3Trace applies a distinction strategy: the Scope Agent extracts
`reported' values for assets with direct disclosures from the evidence
located by the Search Agent, corresponding to Reported Scope Info in
the pipeline, and applies the deterministic estimation of
Appendix~\ref{sec:formulas} for the remainder, corresponding to the
pipeline's Estimation Process. The Verify Agent validates, but never
computes, each derived value through semantic validation and evidence
checking against the asset inventory (the Asset Pool), and all derived
values are labeled with methods, sources, and quality indicators
including extraction confidence scores, supporting the pipeline's
Verification Process.

A key contribution here is semantic relevance validation: the Verify Agent uses semantic similarity models to confirm that estimation rules are only applied where the asset’s attributes described in text have a valid substitution relationship with organizational disclosures, integrating Activity Data and Emission Factor as shown in the pipeline. For example, if an ESG report mentions "office buildings in London" and the asset inventory includes London office buildings, the semantic model verifies the overlap to justify estimation. This constraint, enforced by the Verify Agent’s semantic modules, avoids speculative inferences and maintains alignment with the framework’s evidence-based goal, which is consistent with the pipeline’s emphasis on evidence grounding. Detailed derivation logic, calculation formulas, and factor sources are provided in Appendix~\ref{sec:formulas}.

\subsubsection{Alignment and Consistency Diagnostics}

In the final stage, Scope3Trace conducts conditional matching between building and organizational-level data for diagnostic analysis, which corresponds to the pipeline’s Verification Process, including Consistency Value Check and Logic Check. All three agents collaborate via semantic technologies: the Scope Agent provides semantic-enhanced building-level records with semantic labels from the Asset Pool, the Search Agent provides the normalized Reported Scope Info output, and the Verify Agent uses relation extraction and semantic similarity to evaluate boundary compatibility as a validation step without attempting reconciliation or inference, generating the Evaluation Report as shown in the pipeline.

Matching occurs only when attribution evidence is present such as text snippets in reports linking specific buildings to organizational emissions, aligned with Evidence Grounding Rules in the pipeline and is treated as partial, strictly enforced by the Verify Agent’s semantic modules to prevent over-interpretation. The system does not enforce data reconciliation but uses semantic analysis to determine whether building-level data is applicable semantic alignment between asset and disclosure boundaries, inapplicable semantic mismatch, or lacks alignment evidence no relevant text snippets. These diagnostics, powered by semantic technologies, highlight coverage gaps and boundary limitations, ensuring the system’s output Scope3Trace is transparent, interpretable, and aligned with the pipeline’s overall workflow and the framework’s goal of evidence-based Scope 3 analysis. Additionally, satellite imagery included in the pipeline is used as supplementary contextual information for visual interpretation, supporting asset-level validation without direct involvement in emissions estimation.

\section{Experiments}

\subsection{Extraction Evaluation}
\label{sec:extractionevaluation}

We evaluate the system on a manually annotated gold set covering Scope~1/2/3 totals, Scope~3 categories, reporting year, and page-level evidence (Appendix~\ref{sec:annotation_instructions}), reporting value-level F1 and exact match (EM) for Scope~1/2/3 totals and category F1 for Scope~3 breakdowns. Annotation guidelines were finalized from the GHG Protocol Corporate
Standard before annotation, and the 200-report gold set was frozen
thereafter; all pipeline rules, prompts, and thresholds were developed
exclusively on a disjoint 20-PDF development set, with the gold set used
only for the final evaluation. See Appendix~\ref{app:goldset_validation}
for the full timeline and three independent validation checks, including
an out-of-distribution test (Scope F1 $= 0.98$, EM $= 0.93$). We compare Scope3Trace against a rule-based baseline, two LLM-only regimes (chunked per-page calls on OCR text and a single-call regime that uploads the native PDF) across three model families, a retrieval-augmented pipeline (\citealp{Beck2025}-style), three matched multi-agent frameworks, and ablations of its own components.

As shown in Table~\ref{tab:extraction_results}, Scope3Trace achieves near-perfect performance, significantly outperforming both baselines and standalone LLMs. Ablation results show that removing key components, particularly table reconstruction and evidence grounding, leads to substantial performance drops; page localization, by contrast, mainly affects efficiency rather than accuracy. The ``\textsc{-style}'' rows (\citealp{Beck2025}-style, AutoGen-style, CrewAI-style, MetaGPT-style) denote pipelines that we re-implemented following the design ideas of the corresponding systems on our extraction task, rather than direct invocations of their public code, so that all baselines share the same underlying LLM, prompts, and gold set as Scope3Trace.

\definecolor{groupbg}{HTML}{ECECEC}
\definecolor{winnerbg}{HTML}{FFF3C4}

\begin{table*}[t]
\centering
\small
\setlength{\tabcolsep}{6pt}
\renewcommand{\arraystretch}{1.12}
\begin{tabular}{p{5.2cm}ccccc}
\toprule
\textbf{Method / Model} & \textbf{Scope1 F1} & \textbf{Scope2 F1} & \textbf{Scope3 F1} & \textbf{Category F1} & \textbf{EM} \\
\midrule
\rowcolor{groupbg}\multicolumn{6}{l}{\emph{Non-LLM baseline}} \\
Rule-only & 0.86$\pm$0.00 & 0.86$\pm$0.00 & 0.86$\pm$0.00 & 0.51$\pm$0.01 & 0.45$\pm$0.01 \\
\midrule
\rowcolor{groupbg}\multicolumn{6}{l}{\emph{Chunked LLM (OCR text, per-page calls)}} \\
GPT-4o & 0.84$\pm$0.02 & 0.83$\pm$0.02 & 0.84$\pm$0.02 & 0.70$\pm$0.03 & 0.61$\pm$0.03 \\
Gemini-2.0 & 0.82$\pm$0.03 & 0.81$\pm$0.03 & 0.82$\pm$0.03 & 0.67$\pm$0.04 & 0.58$\pm$0.03 \\
DeepSeek-V3 & 0.80$\pm$0.03 & 0.79$\pm$0.03 & 0.80$\pm$0.03 & 0.64$\pm$0.04 & 0.55$\pm$0.04 \\
\midrule
\rowcolor{groupbg}\multicolumn{6}{l}{\emph{Single-call LLM (native PDF, one call)}} \\
GPT-4o & 0.87$\pm$0.02 & 0.86$\pm$0.02 & 0.87$\pm$0.02 & 0.79$\pm$0.03 & 0.72$\pm$0.03 \\
Gemini-2.0 & 0.86$\pm$0.02 & 0.85$\pm$0.02 & 0.86$\pm$0.02 & 0.77$\pm$0.03 & 0.70$\pm$0.03 \\
DeepSeek-V3 & 0.85$\pm$0.03 & 0.84$\pm$0.03 & 0.85$\pm$0.03 & 0.75$\pm$0.03 & 0.68$\pm$0.02 \\
\midrule
\rowcolor{groupbg}\multicolumn{6}{l}{\emph{Existing LLM pipelines}} \\
\citep{Beck2025}-style 
& 0.88$\pm$0.02 
& 0.87$\pm$0.02 
& 0.88$\pm$0.02 
& 0.78$\pm$0.03 
& 0.71$\pm$0.03 \\
AutoGen-style 
& 0.82$\pm$0.04 
& 0.81$\pm$0.04 
& 0.82$\pm$0.04 
& 0.72$\pm$0.05 
& 0.48$\pm$0.05 \\

CrewAI-style 
& 0.84$\pm$0.03 
& 0.83$\pm$0.03 
& 0.84$\pm$0.03 
& 0.75$\pm$0.04 
& 0.54$\pm$0.04 \\

MetaGPT-style 
& 0.86$\pm$0.03 
& 0.85$\pm$0.03 
& 0.86$\pm$0.03 
& 0.77$\pm$0.04 
& 0.59$\pm$0.04 \\
\midrule
\rowcolor{groupbg}\multicolumn{6}{l}{\emph{Ablations (full system minus one component)}} \\
w/o page localization & 0.95$\pm$0.01 & 0.94$\pm$0.01 & 0.95$\pm$0.01 & 0.93$\pm$0.02 & 0.88$\pm$0.02 \\
w/o table reconstruction & 0.89$\pm$0.02 & 0.88$\pm$0.02 & 0.89$\pm$0.02 & 0.82$\pm$0.03 & 0.75$\pm$0.03 \\
w/o evidence grounding & 0.90$\pm$0.02 & 0.89$\pm$0.02 & 0.88$\pm$0.02 & 0.84$\pm$0.03 & 0.78$\pm$0.02 \\
\midrule
\rowcolor{winnerbg}\textbf{Scope3Trace (GPT-4o)} & \textbf{0.99$\pm$0.004} & \textbf{0.98$\pm$0.006} & \textbf{0.99$\pm$0.004} & \textbf{0.97$\pm$0.008} & \textbf{0.96$\pm$0.01} \\
\bottomrule
\end{tabular}
\caption{Extraction performance on the gold evaluation set. Scope3Trace is compared against a rule-only baseline, two LLM-only regimes (chunked per-page vs.\ single-call on the native PDF) across three model families, a retrieval-augmented LLM pipeline (\citealp{Beck2025}-style), three matched multi-agent frameworks, and ablations of its own components. For aggregated rows we report a single macro Scope~F1 spanning the S1/S2/S3 columns. Results are mean $\pm$ standard deviation over 3 runs.}
\label{tab:extraction_results}
\end{table*}

\subsection{Comparison with Existing Extraction Pipelines}

Compared with the LLM-based benchmark of \citet{Beck2025}, which combines LLM extraction with human validation and provides only \emph{document-level} evidence references, Scope3Trace is a fully automated pipeline evaluated by standard IE metrics (EM and F1 for Scope~1/2/3 and category-level extraction) and provides \emph{page-level} evidence anchoring, linking every extracted value to the specific page containing the disclosure. A side-by-side accuracy comparison against Rule-only, LLM-only, retrieval-augmented (\citealp{Beck2025}-style), and matched multi-agent baselines is included directly in Table~\ref{tab:extraction_results}.

\subsection{Dataset Utility for Prediction}

On a building-level Scope 3 prediction task with multimodal inputs
(building metadata and satellite imagery), a multimodal MLP using the full
Scope3Trace feature set achieves $R^2 = 0.46$ for total Scope 3 and macro
$R^2 = 0.54$ for category-level prediction, improving over single-modality
variants under the same feature setting. For context, the
targets are intrinsically noisy: 96.7\% are proxy-derived
(Table~\ref{tab:provenance}), and commercial emissions-data vendors agree
with one another on Scope 3 at correlations as low as 0.22
($R^2 \approx 0.05$) \citep{busch2022}. The utility claim rests on the
internal ablation: minimal features (floor area and coordinates) yield
$R^2 \le 0.04$ for total and $\le 0.11$ for category-level prediction,
using deliberately simple models (a frozen ResNet-50 with an MLP head) so
that the measured signal reflects the data rather than the model.
Satellite imagery consistently provides complementary signal on top of
tabular features; full setup and results are in
Appendix~\ref{app:prediction_benchmark} and~\ref{app:prediction_results}.

\subsection{Discussion}
Two findings generalize beyond this task. First, under fully matched
conditions, free-form multi-agent orchestration \emph{hurts} extraction:
EM 0.48--0.59 vs.\ 0.72 for a single call, confirmed with official
libraries, and CoT prompting likewise degrades EM to 0.67
(Appendix~\ref{app:autogen}). Second, the rule--LLM boundary, not the
components, determines reliability: four backbones stay within 0.02
Scope F1 and a ${\sim}16\times$ cheaper backbone retains 0.94 category
F1 (Appendix~\ref{app:backbone}), while ablating rule-bounded components
costs up to 0.21 EM (Table~\ref{tab:extraction_results}).

\section{Dataset Analysis and Evaluation}

\subsection{Dataset Coverage and Boundary Applicability}

Scope3Trace contains \textbf{54{,}361 building-year records} (covering 4{,}300+ unique assets, with 17{,}638 non-residential records carrying complete Scope~1/2/3 attribution) and organization-level records across multiple countries in Europe, Australia, and Asia, spanning reporting years 2011--2026, as illustrated in Figure~\ref{fig:geographic_coverage}. Detailed statistics are provided in Appendix~\ref{sec:dataset_summary_tables}. 
The dataset shows broad geographic diversity, with Australia and Europe
contributing the majority of records. The 72.1\% Australian share
reflects data existence rather than selection: NABERS/NGER are the only
nationally unified, audited building inventories, while the organization
level spans multiple regions with 1,300+ EU records
(Table~\ref{tab:org_report_characteristics}).

Alignment coverage remains limited due to structural differences between data sources. Organization-level reports typically provide aggregated emissions without asset-level attribution, while many public building datasets lack explicit ownership linkage. Consequently, coverage diagnostics are treated as contextual metadata rather than quality metrics.

Figure~\ref{fig:scope3_category_availability} illustrates building-level availability
across major Scope 3 categories, with Category 2 and 13 showing the
highest coverage. Restricting building-level modeling to these two
categories is a methodological safeguard: only they have physically
verifiable links to asset attributes (Appendix~\ref{sec:why});
other categories appear only when extracted verbatim
(Table~\ref{tab:provenance}), while the organization level covers all 15
categories. Additional alignment details are provided in
Appendix~\ref{appendix:dataset_schema}.

\begin{figure}[t]
\centering
\includegraphics[width=0.9\linewidth]{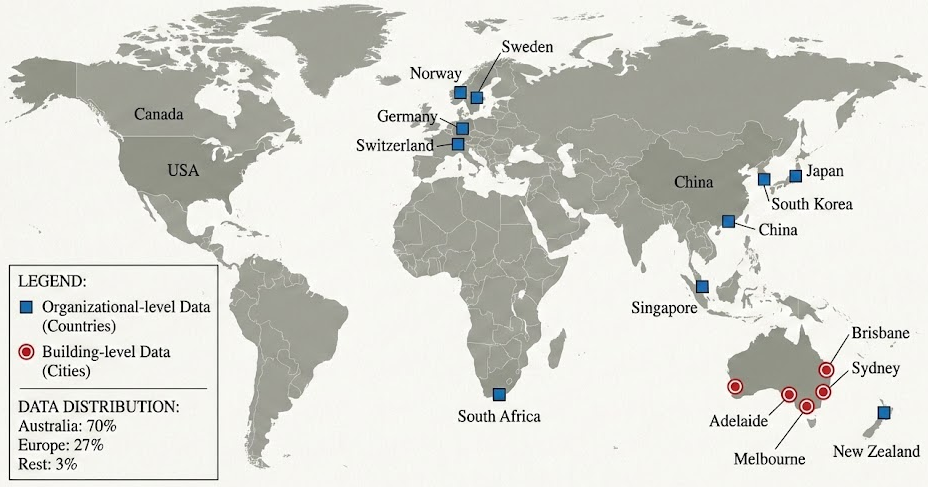}
\caption{Geographic coverage of the Scope3Trace dataset.}
\label{fig:geographic_coverage}
\end{figure}

\subsection{Value Provenance and Confidence Characteristics}

Within the non-residential subset ($N = 17{,}638$) where full Scope~3 estimation is feasible, only 3.28\% of values are directly reported at the building level; the rest are estimated from public energy inventories and emission factors, reflecting the limited state of building-level Scope~3 reporting. Reported values carry high confidence and estimated values medium confidence, the latter expose implicit assumptions and coverage gaps rather than replace reported emissions. Every reported value is additionally anchored to the source page, providing finer-grained provenance than the document-level references common in existing emissions datasets.

\subsection{Organization-Level Disclosure Characteristics}

Scope3Trace also includes organization-level Scope~3 disclosures collected from sustainability and climate reports across industries including chemicals, finance, telecommunications, mining, manufacturing, consumer goods, and technology.

These disclosures span multiple reporting years and jurisdictions. Rather than serving as ground truth for building-level modeling, they function as contextual reference data and control totals, providing insight into reporting boundaries, disclosure completeness, and category coverage.

\paragraph{Cross-level Entity Linking.} Among the unique owning companies of the $\sim$580 building records with directly reported per-building Scope~3 disclosures,  96\% are successfully linked to corresponding organization-level disclosures in Scope3Trace (either directly to company-level reports, or as single-asset SPVs whose asset-level Product Disclosure Statement serves as their complete disclosure). The remaining 4\% are small private owners without public disclosures or SPV/trustee entities whose legal names diverge from parent brands (Appendix~\ref{app:error_analysis}). This demonstrates that within the structurally feasible subset, evidence-grounded cross-level alignment is reliably achieved.

\section{Conclusion}

This paper presents a multimodal dataset and data system designed to support interpretable and coverage-aware analysis of Scope 3 greenhouse gas emissions. By integrating organization-level disclosures with building-level physical inventories under explicit provenance and boundary assumptions, Scope3Trace enables structured examination of how reported emissions relate to underlying assets, activity proxies, and data availability constraints, the dataset emphasizes transparency of reporting assumptions, applicability diagnostics, and reproducible data construction from public data. Our analysis highlights both the opportunities and structural limitations of operationalization of Scope 3 information in real-world settings.

\section*{Limitations}

(1) The alignment between organization-level disclosures and building-level inventories is incomplete: corporate reports seldom disclose building-level emissions, and many building records lack clear ownership attribution. Thus, coverage diagnostics indicate data applicability and evidence availability, rather than reporting completeness.

(2) Building-level Scope 3 representations rely on proxy-based signals from activity data and public emission factors, so they are structured approximations rather than ground-truth measurements.

(3) Geographic and sectoral coverage is uneven: building-level data concentrates on publicly available urban inventories, while organization-level disclosures are drawn from a heterogeneous set of companies across jurisdictions, reflecting data availability.

Specifically, we plan to: (i) strengthen entity linking via
corporate-registry and beneficial-ownership graphs, addressing SPV and
trustee entities whose legal names diverge from parent brands (Appendix \ref{app:error_analysis}); (ii) extend coverage to additional Scope~3 categories via spend-based proxies and supply-chain data; and (iii) augment categorical confidence indicators with \emph{quantitative uncertainty intervals}.

\section*{Acknowledgments}

This research is supported by the ARC Training Centre for Whole Life Design of Carbon Neutral Infrastructure (IC230100015).

\bibliography{custom}

@article{Beck2025,
  author    = {Jacob Beck and Anna Steinberg and Andreas Dimmelmeier and
               Laia Domenech Burin and Emily Kormanyos and Maurice Fehr and
               Malte Schierholz},
  title     = {Addressing Data Gaps in Sustainability Reporting: A Benchmark Dataset for Greenhouse Gas Emission Extraction},
  journal   = {Scientific Data},
  volume    = {12},
  pages     = {1497},
  year      = {2025},
  publisher = {Springer Nature},
  doi       = {10.1038/s41597-025-05664-8}
}

@inproceedings{Guo2025ExioNAICS,
  author    = {Yanming Guo and Jin Ma and Qiao Xiao and Kevin Credit},
  title     = {ExioNAICS: Enterprises Level Emission Estimation Dataset with Large Language Models},
  booktitle = {Proceedings of the ICLR 2025 Workshop on Tackling Climate Change with Machine Learning},
  year      = {2025},
  url       = {https://huggingface.co/datasets/Yvnminc/ExioNAICS}
}

@inproceedings{Mishra2024Statements,
  author    = {Lokesh Mishra and Sohayl Dhibi and Yusik Kim and Cesar Berrospi Ramis and Shubham Gupta and Michele Dolfi and Peter Staar},
  title     = {Statements: Universal Information Extraction from Tables with Large Language Models for ESG KPIs},
  booktitle = {Proceedings of the 1st Workshop on Natural Language Processing Meets Climate Change (ClimateNLP 2024)},
  year      = {2024},
  url       = {https://aclanthology.org/2024.climatenlp-1.15/}
}

@article{Ding2025EulerESG,
  title     = {EulerESG: Automating ESG Disclosure Analysis with LLMs},
  author    = {Yi Ding and Xushuo Tang and Zhengyi Yang and Wenqian Zhang and Simin Wu and Yuxin Huang and Lingjing Lan and Weiyuan Li and Yin Chen and Mingchen Ju and Wenke Yang and Thong Hoang and Mykhailo Klymenko and Xiwei Zu and Wenjie Zhang},
  journal   = {arXiv preprint},
  year      = {2025},
  archivePrefix = {arXiv},
  eprint    = {2511.21712},
  primaryClass = {cs.CL},
  url       = {https://arxiv.org/abs/2511.21712}
}

@article{Hertwich2018Scope3,
  author  = {Edgar G. Hertwich and Richard Wood},
  title   = {The Growing Importance of Scope 3 Greenhouse Gas Emissions from Industry},
  journal = {Environmental Research Letters},
  volume  = {13},
  number  = {10},
  pages   = {104013},
  year    = {2018},
  doi     = {10.1088/1748-9326/aae19a}
}

@inproceedings{Dumit2024Atlas,
  author    = {Andrew Dumit and Krishna Rao and Travis Kwee and
               Varsha Gopalakrishnan and Katherine Tsai and Sangwon Suh},
  title     = {Atlas: A Spend Classification Benchmark for Estimating Scope 3 Carbon Emissions},
  booktitle = {NeurIPS 2024 Workshop on Tackling Climate Change with Machine Learning},
  year      = {2024},
  url       = {https://www.climatechange.ai/papers/neurips2024/70}
}

@misc{sternberg2022cop27,
  author  = {Sternberg, Jeff},
  title   = {At Finance Day, exploring climate risk analysis and ESG entity extraction},
  year    = {2022},
  month   = nov,
  url     = {https://cloud.google.com/blog/topics/sustainability/cop27-climate-risk-analysis-and-esg-entity-extraction},
  note    = {Google Cloud Blog, Sustainability Topics},
}

@inproceedings{dimmelmeier-etal-2024-informing,
    title = {Informing climate risk analysis using textual information - A research agenda},
    author = {Dimmelmeier, Andreas and Doll, Hendrik and Schierholz, Malte and Kormanyos, Emily and Fehr, Maurice and Ma, Bolei and Beck, Jacob and Fraser, Alexander and Kreuter, Frauke},
    editor = {Stammbach, Dominik and Ni, Jingwei and Schimanski, Tobias and Dutia, Kalyan and Singh, Alok and Bingler, Julia and Christiaen, Christophe and Kushwaha, Neetu and Muccione, Veruska and Vaghefi, Saeid A. and Leippold, Markus},
    booktitle = {Proceedings of the 1st Workshop on Natural Language Processing Meets Climate Change (ClimateNLP 2024)},
    month = {Aug},
    year = {2024},
    address = {Bangkok, Thailand},
    publisher = {Association for Computational Linguistics},
    pages = {12--26},
    url = {https://aclanthology.org/2024.climatenlp-1.2/},
    doi = {10.18653/v1/2024.climatenlp-1.2}
}

@article{Yap2025BuildingCarbon,
  author    = {Winston Yap and Abraham Noah Wu and Clayton Miller and Filip Biljecki},
  title     = {Revealing building operating carbon dynamics for multiple cities},
  journal   = {Nature Sustainability},
  year      = {2025},
  doi       = {10.1038/s41893-025-01615-8},
  note      = {Published online 15 August 2025},
  url       = {https://doi.org/10.1038/s41893-025-01615-8}
}

@misc{GHGProtocol2011,
  author = {{Greenhouse Gas Protocol}},
  title = {Corporate Value Chain (Scope 3) Accounting and Reporting Standard},
  year = {2011},
  publisher = {World Resources Institute},
  url = {https://ghgprotocol.org}
}

@article{Lenzen2012MRIO,
  author  = {Lenzen, Manfred and Moran, Daniel and Kanemoto, Keiichiro 
             and Foran, Barney and Lobefaro, Lara and Geschke, Arne},
  title   = {International trade drives biodiversity threats in developing nations},
  journal = {Nature},
  volume  = {486},
  number  = {7401},
  pages   = {109--112},
  year    = {2012},
  doi     = {10.1038/nature11145}
}

@article{traub2025carbon,
  author  = {Traub, Jos{\'e} and Morillas, Carlos and Gil, Rodrigo and {\'A}lvarez, Sergio and Mart{\'i}nez, Sara},
  title   = {Evaluating Corporate Carbon Emissions Reporting: Assessing Transparency and Completeness with the Carbon Integrity Index},
  journal = {Sustainability},
  year    = {2025},
  volume  = {17},
  number  = {17},
  pages   = {7628},
  doi     = {10.3390/su17177628},
  url     = {https://doi.org/10.3390/su17177628}
}

@misc{ghgprotocol2019scope3,
  title={You Too Can Master Value Chain Emissions},
  author={{Greenhouse Gas Protocol}},
  year={2019},
  howpublished={\url{https://ghgprotocol.org/blog/you-too-can-master-value-chain-emissions}}
}

@article{gao2023retrieval,
  title={Retrieval-augmented generation for large language models: A survey},
  author={Gao, Yunfan and Xiong, Yun and Gao, Xinyu and Jia, Kangxiang and Pan, Jinliu and Bi, Yuxi and Dai, Yixin and Sun, Jiawei and Wang, Haofen and Wang, Haofen and others},
  journal={arXiv preprint arXiv:2312.10997},
  volume={2},
  number={1},
  pages={32},
  year={2023}
}

@article{achiam2023gpt,
  title={Gpt-4 technical report},
  author={Achiam, Josh and Adler, Steven and Agarwal, Sandhini and Ahmad, Lama and Akkaya, Ilge and Aleman, Florencia Leoni and Almeida, Diogo and Altenschmidt, Janko and Altman, Sam and Anadkat, Shyamal and others},
  journal={arXiv preprint arXiv:2303.08774},
  year={2023}
}

@article{luu2025mandatory,
  author  = {Nguyen H. Luu and Chau Le and Hiep N. Luu and Dung T. K. Nguyen},
  title   = {Does Mandatory Greenhouse Gas Emissions Reporting Program Deter Corporate Greenwashing?},
  journal = {Journal of Environmental Management},
  volume  = {373},
  pages   = {123740},
  year    = {2025},
  doi     = {10.1016/j.jenvman.2024.123740},
  url     = {https://doi.org/10.1016/j.jenvman.2024.123740}
}

@article{li2025carbon,
  author  = {Mingchen Li and Peigong Li and Wanwan Zhu},
  title   = {Greenhouse Gas Performance and Disclosure--New Global Evidence},
  journal = {International Review of Financial Analysis},
  year    = {2025},
  doi     = {10.1016/j.irfa.2025.104455},
  url     = {https://doi.org/10.1016/j.irfa.2025.104455}
}

@article{wu2023autogen,
  title={AutoGen: Enabling Next-Gen LLM Applications via Multi-Agent Conversation},
  author={Wu, Qingyun and Bansal, Gagan and Zhang, Jieyu and Wu, Yiran and Li, Beibin and Zhu, Erkang and Jiang, Li and Zhang, Xiaoyun and Zhang, Shaokun and Liu, Jiale and Awadallah, Ahmed Hassan and White, Ryen W. and Burger, Doug and Wang, Chi},
  journal={arXiv preprint arXiv:2308.08155},
  year={2023}
}

@misc{nga2025,
  author       = {{Department of Climate Change, Energy, the Environment and Water}},
  title        = {Australian National Greenhouse Accounts Factors: 2025},
  year         = {2025},
  howpublished = {\url{https://www.dcceew.gov.au/climate-change/publications/national-greenhouse-accounts-factors-2025}},
  note         = {Accessed: 2026-07-12}
}

@misc{nga2019,
  author       = {{Department of the Environment and Energy}},
  title        = {Australian National Greenhouse Accounts Factors: 2019},
  year         = {2019},
  note         = {Accessed: 2026-07-12}
}

@misc{com_energy,
  author = {{City of Melbourne}},
  title  = {Property Level Energy Consumption (Modelled on Building Attributes): Baseline 2011 and Business-as-Usual Projections 2016--2026},
  year   = {2014},
  note   = {City of Melbourne Open Data Portal. Accessed: 2025-12-08}
}

@article{busch2022,
  author  = {Busch, Timo and Johnson, Matthew and Pioch, Thomas},
  title   = {Corporate Carbon Performance Data: Quo Vadis?},
  journal = {Journal of Industrial Ecology},
  volume  = {26},
  number  = {1},
  pages   = {350--363},
  year    = {2022},
  doi     = {10.1111/jiec.13008}
}

@misc{crewai,
  author       = {Moura, Jo{\~a}o},
  title        = {CrewAI: Framework for Orchestrating Role-Playing, Autonomous AI Agents},
  year         = {2023},
  howpublished = {\url{https://github.com/crewAIInc/crewAI}},
  note         = {Version 1.15.2}
}

@misc{metagpt,
  title        = {MetaGPT: Meta Programming for A Multi-Agent Collaborative Framework},
  author       = {Hong, Sirui and Zhuge, Mingchen and Chen, Jonathan and Zheng, Xiawu and Cheng, Yuheng and Zhang, Ceyao and Wang, Jinlin and Wang, Zili and Yau, Steven Ka Shing and Lin, Zijuan and Zhou, Liyang and Ran, Chenyu and Xiao, Lingfeng and Wu, Chenglin and Schmidhuber, J{\"u}rgen},
  year         = {2023},
  eprint       = {2308.00352},
  archivePrefix = {arXiv},
  primaryClass = {cs.AI}
}

@inproceedings{xue2026supervised,
  title={Why supervised fine-tuning fails to learn: A systematic study of incomplete learning in large language models},
  author={Xue, Chao and Wang, Yao and Liu, Mengqiao and Liang, Di and Han, Xingsheng and Liu, Peiyang and Wu, Xianjie and Lu, Chenyao and Jiang, Lei and Lu, Yu and others},
  booktitle={Proceedings of the 64th Annual Meeting of the Association for Computational Linguistics (Volume 1: Long Papers)},
  pages={30186--30213},
  year={2026}
}

@inproceedings{xue2026reason,
  title={Reason only when needed: Efficient generative reward modeling via model-internal uncertainty},
  author={Xue, Chao and Wang, Yao and Liu, Mengqiao and Liang, Di and Han, Xingsheng and Liu, Peiyang and Wu, Xianjie and Lu, Chenyao and Jiang, Lei and Lu, Yu and others},
  booktitle={Findings of the Association for Computational Linguistics: ACL 2026},
  pages={23302--23319},
  year={2026}
}

@inproceedings{xue2024question,
  title={Question calibration and multi-hop modeling for temporal question answering},
  author={Xue, Chao and Liang, Di and Wang, Pengfei and Zhang, Jing},
  booktitle={Proceedings of the AAAI Conference on Artificial Intelligence},
  volume={38},
  number={17},
  pages={19332--19340},
  year={2024}
}

@inproceedings{xue2023dual,
  title={Dual path modeling for semantic matching by perceiving subtle conflicts},
author={Xue, Chao and Liang, Di and Wang, Sirui and Zhang, Jing and Wu, Wei},
  booktitle={ICASSP 2023-2023 IEEE International Conference on Acoustics, Speech and Signal Processing (ICASSP)},
  pages={1--5},
  year={2023},
  organization={IEEE}
}

@inproceedings{xue2025structcoh,
  title={Structcoh: Structured contrastive learning for context-aware text semantic matching},
  author={Xue, Chao and Gao, Ziyuan},
  booktitle={Pacific Rim International Conference on Artificial Intelligence},
  pages={300--315},
  year={2025},
  organization={Springer}
}

@inproceedings{li2026draftrl,
  author    = {Li, Yuanhao and Liu, Mingshan and Wang, Hongbo and Zhang, Yiding and Ma, Yifei and Tan, Wei},
  title     = {{DRAFT-RL}: Multi-Agent Chain-of-Draft Reasoning for Reinforcement Learning-Enhanced {LLMs}},
  booktitle = {Proceedings of the AAAI Conference on Artificial Intelligence, Vol. 40, No. 35},
  year      = {2026},
  pages     = {29530--29537},
  publisher = {Association for the Advancement of Artificial Intelligence},
  doi       = {10.1609/aaai.v40i35.40195},
  url       = {https://doi.org/10.1609/aaai.v40i35.40195}
}

@inproceedings{li2026boostapr,
  author    = {Li, Yuanhao and Wang, Hongbo and Shang, Xiaotang and Tang, Xunzhu and Cao, Yiming and Chen, Xuhong},
  title     = {{BoostAPR}: Boosting Automated Program Repair via Execution-Grounded Reinforcement Learning with Dual Reward Models},
  booktitle = {Proceedings of the Forty-Third International Conference on Machine Learning},
  year      = {2026},
  series    = {Proceedings of Machine Learning Research},
  volume    = {306},
  publisher = {PMLR},
  url       = {https://openreview.net/forum?id=J2TKCPUI3E}
}

@inproceedings{tan2026evocurl,
  author    = {Tan, Wei and Li, Yuanhao and Liang, Wenkai},
  title     = {{Evo-CuRL}: Curriculum-Aware Reinforcement Learning over Code Lineage Graphs for Software Engineering Reasoning},
  booktitle = {Proceedings of the 2026 International Conference on Multimedia Retrieval},
  year      = {2026},
  pages     = {1327--1335},
  publisher = {ACM},
  doi       = {10.1145/3805622.3810701},
  url       = {https://doi.org/10.1145/3805622.3810701}
}

@inproceedings{li2026retrieval,
  author    = {Li, Yuanhao and Wang, Hongbo and Chen, Xuhong and Cao, Yiming},
  title     = {Retrieval-Augmented Multi-Agent Multimodal Framework for Fake News Detection},
  booktitle = {ICASSP 2026 -- 2026 IEEE International Conference on Acoustics, Speech and Signal Processing (ICASSP)},
  year      = {2026},
  pages     = {14142--14146},
  publisher = {IEEE},
  doi       = {10.1109/ICASSP55912.2026.11462174},
  url       = {https://doi.org/10.1109/ICASSP55912.2026.11462174}
}

@misc{zhu2026fcpragfusioncontrollerparametricretrievalaugmented,
      title={FCPRAG: Fusion-Controller Parametric Retrieval-Augmented Generation for Stable Multi-Passage LoRA Injection}, 
      author={Jinchang Zhu and Jindong Li and Yi Ding and Xiaojian Nie and Rong Fu and Shuangyong Song and Haowei He and Menglin Yang},
      year={2026},
      eprint={2608.21750},
      archivePrefix={arXiv},
      primaryClass={cs.CL},
      url={https://arxiv.org/abs/2608.21750}, 
}

@inproceedings{zhu2026hela,
  title={HeLa-Mem: Hebbian Learning and Associative Memory for LLM Agents},
  author={Zhu, Jinchang and Li, Jindong and Zhang, Cheng and Liu, Jiahong and Yang, Menglin},
  booktitle={Proceedings of the 64th Annual Meeting of the Association for Computational Linguistics (Volume 1: Long Papers)},
  pages={13757--13769},
  year={2026}
}

@misc{zhu2026learninglookagainlossgap,
      title={Learning to Look Again: Loss-Gap Supervision for Free-form Crop Routing in Vision-Language Models}, 
      author={Jinchang Zhu and Rong Fu and Yi Ding and Chenghao Wu and Ying Liu and Menglin Yang},
      year={2026},
      eprint={2608.21762},
      archivePrefix={arXiv},
      primaryClass={cs.CV},
      url={https://arxiv.org/abs/2608.21762}, 
}

@misc{zhu2026doeslongcontextsupervisionactually,
      title={Where Does Long-Context Supervision Actually Go? Effective-Context Exposure Balancing}, 
      author={Jinchang Zhu and Jindong Li and Chengyu Zou and Rong Fu and Chao Wang and Haowei He and Menglin Yang},
      year={2026},
      eprint={2605.10544},
      archivePrefix={arXiv},
      primaryClass={cs.CL},
      url={https://arxiv.org/abs/2605.10544}, 
}

@misc{lai2026biasscopeautomateddetectionbias,
      title={BiasScope: Towards Automated Detection of Bias in LLM-as-a-Judge Evaluation},
      author={Peng Lai and Zhihao Ou and Yong Wang and Longyue Wang and Jian Yang and Yun Chen and Guanhua Chen},
      year={2026},
      eprint={2602.09383},
      archivePrefix={arXiv},
      primaryClass={cs.CL},
      url={https://arxiv.org/abs/2602.09383}, 
}

@misc{zhu2026anchoredsupervisedfinetuning,
      title={Anchored Supervised Fine-Tuning},
      author={He Zhu and Junyou Su and Peng Lai and Ren Ma and Wenjia Zhang and Linyi Yang and Guanhua Chen},
      year={2026},
      eprint={2509.23753},
      archivePrefix={arXiv},
      primaryClass={cs.LG},
      url={https://arxiv.org/abs/2509.23753}, 
}

@misc{lai2025surfaceenhancingllmasajudgealignment,
      title={Beyond the Surface: Enhancing LLM-as-a-Judge Alignment with Human via Internal Representations},
      author={Peng Lai and Jianjie Zheng and Sijie Cheng and Yun Chen and Peng Li and Yang Liu and Guanhua Chen},
      year={2025},
      eprint={2508.03550},
      archivePrefix={arXiv},
      primaryClass={cs.CL},
      url={https://arxiv.org/abs/2508.03550}, 
}

@inproceedings{lai2026unirrm,
title={Uni{RRM}: Unified Reasoning Reward Models Across Languages and Evaluation Paradigms},
author={Peng Lai and Yichao Du and Junchao Wu and Weibo Gao and Linan Yue and Longyue Wang and Weihua Luo and Derek F. Wong and Guanhua Chen},
booktitle={Forty-third International Conference on Machine Learning},
year={2026},
url={https://openreview.net/forum?id=laiK6TlhL2}
}

@misc{lai2026aligndiff,
title={AlignDiff: Exploiting Model-Intrinsic Information for Better Preference Data Selection},
author={Peng Lai and He Zhu and Zhiwen Ruan and Dongdong Zhang and Yun Chen and Peng Li and Furu Wei and Yang Liu and Guanhua Chen},
year={2026},
url={https://openreview.net/forum?id=4fclVIrUg2}
}

@article{sheng2026group,
  title={Group LASSO for multiple change-point detection in a generalized integer-valued autoregressive model: D. Sheng et al.},
  author={Sheng, Danshu and Hua, Jiao and Xu, Jinxian and Wang, Dehui and Jin, Shuilin},
  journal={Statistical Papers},
  volume={67},
  number={5},
  pages={100},
  year={2026},
  publisher={Springer}
}

@article{qi2025over++,
  title={Over++: Generative video compositing for layer interaction effects},
  author={Qi, Luchao and Wu, Jiaye and Choi, Jun Myeong and Phillips, Cary and Sengupta, Roni and Goldman, Dan B},
  journal={arXiv preprint arXiv:2512.19661},
  year={2025}
}

@article{zhao2026deployment,
  title={Do Deployment Constraints Make LLMs Hallucinate Citations? An Empirical Study across Four Models and Five Prompting Regimes},
  author={Zhao, Chen and Tang, Yuan and Qian, Yitian},
  journal={arXiv preprint arXiv:2603.07287},
  year={2026}
}

@inproceedings{lu2026s,
  title={S 3 G: Stock State Space Graph for Enhanced Stock Trend Prediction},
  author={Lu, Yao and Hu, Kaiyi and Zhang, Luyan},
  booktitle={ICASSP 2026-2026 IEEE International Conference on Acoustics, Speech and Signal Processing (ICASSP)},
  pages={4081--4085},
  year={2026},
  organization={IEEE}
}

@article{yuan2026sp,
  title={SP-Mind: An Autonomous Reasoning Agent for Spatial Proteomics Analysis},
  author={Yuan, Yucheng and Ji, Yuanfeng and Li, Zhongxiao and Li, Ruijiang},
  journal={arXiv preprint arXiv:2606.24235},
  year={2026}
}

@article{kuang2025learning,
  title={Learning Game-Playing Agents with Generative Code Optimization},
  author={Kuang, Zhiyi and Rong, Ryan and Yuan, YuCheng and Nie, Allen},
  journal={arXiv preprint arXiv:2508.19506},
  year={2025}
}

@article{nie2026understanding,
  title={Understanding the challenges in iterative generative optimization with LLMs},
  author={Nie, Allen and Daull, Xavier and Kuang, Zhiyi and Akkiraju, Abhinav and Chaudhuri, Anish and Piasevoli, Max and Rong, Ryan and Yuan, YuCheng and Choudhary, Prerit and Xiao, Shannon and others},
  journal={arXiv preprint arXiv:2603.23994},
  year={2026}
}

@misc{yang2026trilens,
  title         = {{TriLens}: Per-Layer Logit-Lens Entropy for White-Box Hallucination Detection},
  author        = {Yang, Bohan and Gong, Yijun and Zhang, Zhi and Zhang, Ge and Xing, Wenpeng and Han, Meng},
  year          = {2026},
  eprint        = {2606.01033},
  archivePrefix = {arXiv},
  primaryClass  = {cs.AI},
  doi           = {10.48550/arXiv.2606.01033},
  url           = {https://arxiv.org/abs/2606.01033}
}

@article{xu2026beyond,
  title={Beyond GRPO and on-policy distillation: An empirical sparse-to-dense reward principle for language-model post-training},
  author={Xu, Yuanda and Sang, Hejian and Zhou, Zhengze and He, Ran and Wang, Zhipeng and Geramifard, Alborz},
  journal={arXiv preprint arXiv:2605.12483},
  year={2026}
}

@article{xu2026tip,
  title={Tip: Token importance in on-policy distillation},
  author={Xu, Yuanda and Sang, Hejian and Zhou, Zhengze and He, Ran and Wang, Zhipeng and Geramifard, Alborz},
  journal={arXiv preprint arXiv:2604.14084},
  year={2026}
}

@article{xu2026paced,
  title={PACED: Distillation and on-policy self-distillation at the frontier of student competence},
  author={Xu, Yuanda and Sang, Hejian and Zhou, Zhengze and He, Ran and Wang, Zhipeng},
  journal={arXiv preprint arXiv:2603.11178},
  year={2026}
}

@article{yu2025test,
  title={Test-time immunization: A universal defense framework against jailbreaks for (multimodal) large language models},
  author={Yu, Yongcan and Wang, Yanbo and He, Ran and Liang, Jian},
  journal={arXiv preprint arXiv:2505.22271},
  year={2025}
}

@article{wang2025comprehensive,
  title={A comprehensive survey on trustworthiness in reasoning with large language models},
  author={Wang, Yanbo and Yu, Yongcan and Liang, Jian and He, Ran},
  journal={arXiv preprint arXiv:2509.03871},
  year={2025}
}

@article{yu2025reassessing,
  title={Reassessing the role of supervised fine-tuning: an empirical study in VLM reasoning},
  author={Yu, Yongcan and He, Lingxiao and Lu, Shuo and Sheng, Lijun and Xu, Yinuo and Wang, Yanbo and Guo, Kuangpu and Cheng, Jianjie and Wang, Meng and Xie, Qianlong and others},
  journal={arXiv preprint arXiv:2512.12690},
  year={2025}
}

@inproceedings{yu2026understanding,
  title={Understanding and mitigating spurious signal amplification in test-time reinforcement learning for math reasoning},
  author={Yu, Yongcan and He, Lingxiao and Liang, Jian and Guo, Kuangpu and Wang, Meng and Xie, Qianlong and Wang, Xingxing and He, Ran},
  booktitle={Findings of the Association for Computational Linguistics: ACL 2026},
  pages={37424--37436},
  year={2026}
}

@misc{du2026possibledefinitebenchmarkevaluating,
      title={Possible or Definite? A Benchmark for Evaluating Diagnostic Uncertainty Preservation in Clinical Text}, 
      author={Hongbo Du and Zixin Lu and Jiaming Qu},
      year={2026},
      eprint={2606.18471},
      archivePrefix={arXiv},
      primaryClass={cs.CL},
      url={https://arxiv.org/abs/2606.18471}, 
}

@INPROCEEDINGS{8615588,
  author={Du, Hongbo and Jouny, Ismail and Yu, Yih-Choung and Wang, Sicheng},
  booktitle={2018 IEEE Signal Processing in Medicine and Biology Symposium (SPMB)}, 
  title={Exploring P300-based Biometric for Individual Identification Based on Convolutional Neural Networks}, 
  year={2018},
  volume={},
  number={},
  pages={1-3},
  doi={10.1109/SPMB.2018.8615588}}

@inproceedings{zhou2026comem,
title={CoMem: Compositional Concept-Graph Memory for Vision--Language Adaptation},
author={Zhou, Heng and Tang, Jing and Zhang, Jusheng and Li, Yanshu and Xiao, Canran and Hou, Liwei and Ke, Zong and Yao, Jiawei},
booktitle={International Conference on Learning Representations},
volume={2026},
pages={20009--20032},
year={2026}
}

@inproceedings{ye2026seeing,
title={Seeing through the Conflict: Transparent Knowledge Conflict Handling in Retrieval-Augmented Generation},
author={Ye, Hua and Chen, Siyuan and Zhong, Ziqi and Xiao, Canran and Zhang, Haoliang and Wu, Yuhan and Shen, Fei},
booktitle={Proceedings of the AAAI Conference on Artificial Intelligence},
volume={40},
number={40},
pages={34423--34431},
year={2026}
}

@inproceedings{lin2026cec,
title={Cec-zero: Zero-supervision character error correction with self-generated rewards},
author={Lin, Zhiming and Zhao, Kai and Zhang, Sophie and Yu, Peilai and Xiao, Canran},
booktitle={Proceedings of the AAAI Conference on Artificial Intelligence},
volume={40},
number={28},
pages={23612--23620},
year={2026}
}

@article{xiao2025multifrequency,
title={A multifrequency data fusion deep learning model for carbon price prediction},
author={Xiao, Canran and Liu, Yongmei},
journal={Journal of Forecasting},
volume={44},
number={2},
pages={436--458},
year={2025},
publisher={Wiley Online Library}
}

@inproceedings{xiao2026points,
title={From points to coalitions: Hierarchical contrastive shapley values for prioritizing data samples},
author={Xiao, Canran and Dou, Jiabao and Lin, Zhiming and Ke, Zong and Hou, Liwei},
booktitle={Proceedings of the AAAI Conference on Artificial Intelligence},
volume={40},
number={19},
pages={15995--16003},
year={2026}
}

@inproceedings{hongmo2026lightweight,
  title={Lightweight Adaptation of BERT for Negation Sentiment Classification},
  author={Hongmo, Tao and Kamata, Hiroyuki},
  booktitle={2026 International Conference on Artificial Intelligence in Information and Communication (ICAIIC)},
  pages={762--766},
  year={2026},
  organization={IEEE}
}

@inproceedings{shao2026continuous,
  title={Continuous Trust in multi-cloud Environments: An Evidence-Driven Security Assurance Method},
  author={Shao, Wanru},
  booktitle={2026 2nd International Conference on Emerging and Intelligent Technologies and Systems (EITS)},
  pages={123--128},
  year={2026},
  organization={IEEE}
}

@article{shao2026design,
  title={Design and Implementation of an Open-Source Security Framework for Cloud Infrastructure},
  author={Shao, Wanru},
  journal={arXiv preprint arXiv:2604.03331},
  year={2026}
}

@INPROCEEDINGS{11460600,
  author={Gao, Yaxin and Lu, Yao and Zhang, Zongfei and Nie, Jiaqi and Yu, Shanqing and Xuan, Qi},
  booktitle={ICASSP 2026 - 2026 IEEE International Conference on Acoustics, Speech and Signal Processing (ICASSP)}, 
  title={DSPC: Dual-Stage Progressive Compression Framework for Efficient Long-Context Reasoning}, 
  year={2026},
  volume={},
  number={},
  pages={19387-19391},
  doi={10.1109/ICASSP55912.2026.11460600}}

@InProceedings{10.1007/978-981-95-4100-3_2,
author="Jiang, Feng
and Zhang, Zongfei
and Xu, Xin",
editor="Taniguchi, Tadahiro
and Leung, Chi Sing Andrew
and Kozuno, Tadashi
and Yoshimoto, Junichiro
and Mahmud, Mufti
and Doborjeh, Maryam
and Doya, Kenji",
title="CMFDNet: Cross-Mamba and Feature Discovery Network for Polyp Segmentation",
booktitle="Neural Information Processing",
year="2026",
publisher="Springer Nature Singapore",
address="Singapore",
pages="17--30",
isbn="978-981-95-4100-3"
}

@InProceedings{10.1007/978-981-95-4384-7_10,
author="Lou, Yuwei
and Hu, Hao
and Ma, Shaocong
and Zhang, Zongfei
and Wang, Liang
and Ge, Jidong
and Tao, Xianping",
editor="Taniguchi, Tadahiro
and Leung, Chi Sing Andrew
and Kozuno, Tadashi
and Yoshimoto, Junichiro
and Mahmud, Mufti
and Doborjeh, Maryam
and Doya, Kenji",
title="DRF: LLM-AGENT Dynamic Reputation Filtering Framework",
booktitle="Neural Information Processing",
year="2026",
publisher="Springer Nature Singapore",
address="Singapore",
pages="127--141",
isbn="978-981-95-4384-7"
}

@article{zhang2025lightweight,
  title={A Lightweight Skeleton-Based Fall Detection Framework Using Multi-Dimensional Attention Mechanisms},
  author={Zhang, Zongfei and Ni, Haoze},
  journal={INNO-PRESS: Journal of Emerging Applied AI},
  volume={1},
  number={9},
  year={2025}
}

@inproceedings{lin2026parameter,
  title={Parameter Importance is Not Static: Evolving Parameter Isolation for Supervised Fine-Tuning},
  author={Lin, Zekai and Xue, Chao and Liang, Di and Han, Xingsheng and Liu, Peiyang and Wu, Xianjie and Jiang, Lei and Lu, Yu and Simons, Bob and Liang, Shuang and others},
  booktitle={Proceedings of the 64th Annual Meeting of the Association for Computational Linguistics (Volume 1: Long Papers)},
  pages={32846--32864},
  year={2026}
}

@inproceedings{wen2026reinforcement,
  title={Reinforcement Learning Enhanced Muti-hop Reasoning for Temporal Knowledge Question Answering},
  author={Wen, Wuzhenghong and Xue, Chao and Pan, Su and Sun, Yuwei and Peng, Minlong},
  booktitle={Proceedings of the AAAI Conference on Artificial Intelligence},
  volume={40},
  number={40},
  pages={33881--33889},
  year={2026}
}

@inproceedings{dai2025seper,
  title={Seper: Measure retrieval utility through the lens of semantic perplexity reduction},
  author={Dai, Lu and Xu, Yijie and Ye, Jinhui and Liu, Hao and Xiong, Hui},
  booktitle={International Conference on Learning Representations},
  volume={2025},
  pages={78977--79007},
  year={2025}
}

@article{dai2026llm,
  title={LLM-Oriented Information Retrieval: A Denoising-First Perspective},
  author={Dai, Lu and Sun, Liang and Cao, Fanpu and Rao, Ziyang and Yang, Cehao and Liu, Hao and Xiong, Hui},
  journal={arXiv preprint arXiv:2605.00505},
  year={2026}
}

@inproceedings{li-etal-2025-sensorllm,
  title = "{S}ensor{LLM}: Aligning Large Language Models with Motion Sensors for Human Activity Recognition",
  author = "Li, Zechen and Deldari, Shohreh and Chen, Linyao and Xue, Hao and Salim, Flora D.",
  editor = "Christodoulopoulos, Christos and Chakraborty, Tanmoy and Rose, Carolyn and Peng, Violet",
  booktitle = "Proceedings of the 2025 Conference on Empirical Methods in Natural Language Processing",
  month = nov,
  year = "2025",
  address = "Suzhou, China",
  publisher = "Association for Computational Linguistics",
  url = "https://aclanthology.org/2025.emnlp-main.19/",
  doi = "10.18653/v1/2025.emnlp-main.19",
  pages = "354--379",
  ISBN = "979-8-89176-332-6",
}

@inproceedings{li-etal-2026-zara,
  title = "{ZARA}: Training-Free Motion Time-Series Reasoning via Evidence-Grounded {LLM} Agents",
  author = "Li, Zechen and Chen, Baiyu and Xue, Hao and Salim, Flora D.",
  editor = "Liakata, Maria and Moreira, Viviane P. and Zhang, Jiajun and Jurgens, David",
  booktitle = "Proceedings of the 64th Annual Meeting of the {A}ssociation for {C}omputational {L}inguistics (Volume 1: Long Papers)",
  month = jul,
  year = "2026",
  address = "San Diego, California, United States",
  publisher = "Association for Computational Linguistics",
  url = "https://aclanthology.org/2026.acl-long.684/",
  doi = "10.18653/v1/2026.acl-long.684",
  pages = "14986--15008",
  ISBN = "979-8-89176-390-6",
}

@article{duan2026ghgbench,
  title={{GHGbench}: A Unified Multi-Entity, Multi-Task Benchmark for Carbon Emission Prediction},
  author={Duan, Yifan and Zheng, Siyuan and Li, Lihuan and Xue, Chao and Salim, Flora},
  journal={arXiv preprint arXiv:2605.13743},
  url={https://arxiv.org/abs/2605.13743},
  year={2026}
}

\FloatBarrier
\clearpage
\appendix

\section{Dataset Schema Overview}
\label{appendix:dataset_schema}

This appendix provides an overview of the dataset schema used in the 
\textbf{Scope3Trace} dataset, which integrates building-level greenhouse gas 
(GHG) emissions data with organization-level carbon disclosure records. 
The schema is designed to support multi-scale carbon accounting research, 
particularly Scope~3 emissions analysis in the built environment.

The dataset aligns with the GHG Protocol reporting framework while supporting
multi-year, multimodal, and cross-organizational analysis.
All source disclosures and extracted records are in English. The dataset documents
corporate entities and physical building assets, and contains no individual-level
or human demographic data.

\subsection{Design Principles}

The schema was developed based on four primary design considerations:

\begin{itemize}
    \item \textbf{GHG Protocol alignment.}
    Scope~1, Scope~2 (location-based and market-based), and Scope~3 category 
    definitions follow established carbon accounting standards.

    \item \textbf{Cross-scale linkage.}
    Building-level asset records are linked to organization-level disclosures 
    using a shared \texttt{org\_id}, enabling hierarchical emissions analysis.

    \item \textbf{Temporal consistency.}
    Each record corresponds to a specific reporting year to enable 
    longitudinal carbon disclosure analysis.

    \item \textbf{Data transparency and provenance.}
    Source type, estimation method, and qualitative confidence indicators are 
    recorded for reproducibility and uncertainty analysis.

    \item \textbf{Multimodal extensibility.}
    Optional metadata such as satellite imagery paths allow integration of 
    remote sensing and geospatial features for AI-driven analysis.
\end{itemize}

\subsection{Building-Level Dataset}

The building-level dataset is stored in CSV format, where each row represents 
a building asset for a specific reporting year. The dataset captures:

\begin{itemize}
    \item asset identification, geolocation, and organization linkage;
    \item building physical attributes (floor area, usage type, construction year);
    \item energy consumption data;
    \item Scope~1 and Scope~2 emissions with methodological metadata;
    \item selected Scope~3 categories, particularly:
    \begin{itemize}
        \item Category 2 (capital goods / embodied carbon),
        \item Category 13 (downstream leased assets).
    \end{itemize}
\end{itemize}

These categories are prioritised because they can be partially inferred from 
asset-level characteristics such as building size, operational energy use, 
and tenant activity.

\paragraph{Full Raw Sample (Building-Level)}
\label{sam:smapleofbuildinglevel}

For reproducibility, a representative raw record is provided below.

\begin{lstlisting}[basicstyle=\ttfamily\footnotesize]
{
  "property_id": 102104,
  "building_name": "Rialto",
  "address_standardized": "525 COLLINS STREET | MELBOURNE | VIC 3000",
  "org_id": "org_33945",
  "report_year": 2024,
  "floor_area_sqm": 5021.93,
  "energy_electricity_kwh": 27089.43,
  "energy_data_source": "reported",

  "scope1": {
    "value_tco2e": 1382.4,
    "source": "reported",
    "evidence": {
      "page": 44,
      "text_evidence": "Sub-total (assets under management) | 1,525 | 1,555 | 73,112 | 76,191 | ..."
    }
  },
  "scope2": {
    "location_based_tco2e": 29.417,
    "market_based_tco2e": 29.417,
    "source": "reported",
    "evidence": {
      "page": 44,
      "text_evidence": "Energy use | Electricity (location-based) | 2,735 | ... | (market-based) | 29.417 | ..."
    }
  },
  "scope3_total": {
    "value_tco2e": 246.1,
    "source": "reported",
    "evidence": {
      "page": 12,
      "text_evidence": "Scope 3 total emissions from PDS"
    }
  },
  "scope3_breakdown": {
    "cat13_leased_assets": {
      "value_tco2e": 246.1,
      "source": "reported",
      "evidence": {
        "page": 12,
        "text_evidence": "Scope 3 total emissions from PDS"
      }
    }
  },
  "image_path": "data/media/satellite/3000/525-COLLINS-STREET-MELBOURNE-VIC-3000__pid102104__1024x1024at2x.png"
}
\end{lstlisting}

\subsection{Organization-Level Dataset}

The organization-level dataset is stored in JSON Lines (JSONL) format, with one 
record per organization per reporting year. Each record includes:

\begin{itemize}
    \item Organizational identifiers and reporting boundary definitions;
    \item Enterprise-wide Scope~1 and Scope~2 totals;
    \item A nested Scope~3 summary object containing total emissions and 
    category-level breakdowns when disclosed.
    \item Restatement metadata: a \texttt{restated} flag and, where
applicable, a \texttt{superseded\_by} link to the superseding
observation.
\end{itemize}

When a later report restates a prior figure, both observations are
retained with their own evidence: the restating record carries
\texttt{restated:\,true}, and the superseded record links forward via
\texttt{superseded\_by} (analysis in Appendix~\ref{app:restatements}).

This structure reflects common sustainability reporting practices while enabling 
fine-grained analysis of Scope~3 disclosure completeness.

\paragraph{Full Raw Sample (Organization-Level)}

A representative organization-level record is provided below.

\begin{lstlisting}[caption={Organization-level carbon disclosure with evidence grounding (Enel, 2024).}]
{
  "org_id": "ORG_ENEL",
  "org_name_standardized": "Enel",
  "report_year": 2024,
  "reporting_boundary": "Operational Control (Assumed)",
  "assurance_level": "Unknown",
  "scope1_tco2e_value": 20209310.0,
  "scope1_provenance": "reported",
  "scope1_evidence": {
    "page": 24,
    "text_evidence": "In 2024, Scope 1 GHG emissions amounted to 20,209,310 tCO2eq."
  },
  "scope2_location_tco2e_value": 3131164.0,
  "scope2_market_tco2e_value": 4898276.0,
  "scope2_provenance": "reported",
  "scope2_evidence": {
    "location_based": {
      "page": 25,
      "text_evidence": "Total Scope 2 (location-based) 3,131,164 tCO2eq."
    },
    "market_based": {
      "page": 25,
      "text_evidence": "Total Scope 2 (market-based) 4,898,276 tCO2eq."
    }
  },
  "scope3_summary": {
    "total_tco2e_value": 46259089.0,
    "provenance": "reported",
    "total_evidence": {
      "page": 26,
      "text_evidence": "In 2024, Scope 3 GHG emissions amounted to 46,259,089 tCO2eq."
    },
    "breakdown": {
      "cat1_purchased_goods_tco2e": 4344969.0,
      "cat2_capital_goods_tco2e": 3830188.0,
      "cat3_fuel_energy_tco2e": 23734668.0,
      "cat4_upstream_transport_tco2e": 7518.0,
      "cat6_business_travel_tco2e": 21243.0,
      "cat7_employee_commuting_tco2e": 36399.0,
      "cat11_use_sold_tco2e": 14284105.0
    },
    "breakdown_evidence": {
      "cat1_purchased_goods_tco2e": {
        "page": 30,
        "text_evidence": "Cat.1 Purchased goods and services 4,344,969 (tons CO2-eq)."
      },
      "cat2_capital_goods_tco2e": {
        "page": 30,
        "text_evidence": "Cat.2 Capital Goods 3,830,188 (tons CO2-eq)."
      },
      "cat3_fuel_energy_tco2e": {
        "page": 30,
        "text_evidence": "Cat.3 Fuel and Energy related activities 23,734,668 (tons CO2-eq)."
      },
      "cat4_upstream_transport_tco2e": {
        "page": 30,
        "text_evidence": "Cat.4 Upstream transportation and distribution 7,518 (tons CO2-eq)."
      },
      "cat6_business_travel_tco2e": {
        "page": 30,
        "text_evidence": "Cat.6 Business travel 21,243 (tons CO2-eq)."
      },
      "cat7_employee_commuting_tco2e": {
        "page": 30,
        "text_evidence": "Cat.7 Commuting 36,399 (tons CO2-eq)."
      },
      "cat11_use_sold_tco2e": {
        "page": 30,
        "text_evidence": "Cat.11 natural gas sold in the retail market 14,284,105 (tons CO2-eq)."
      }
    }
  },
  "restated": false,
  "superseded_by": null,
  "source_document": "ghg-inventory-2024.pdf",
  "schema_version": "v2.2"
}
\end{lstlisting}

\subsection{Cross-Level Consistency Constraints}

To ensure logical consistency:

\begin{itemize}
    \item Aggregated building Scope~1 and Scope~2 emissions should not exceed 
    organization-level totals.

    \item Market-based Scope~2 emissions are typically less than or equal to 
    location-based values.

    \item Building-level Scope~3 coverage may be incomplete due to disclosure 
    limitations and reporting boundaries.
\end{itemize}

\subsection{Geographic Coverage}

The Scope3Trace dataset primarily covers assets and disclosures from Australia
and Europe, reflecting the availability of high-quality building inventories
and publicly accessible sustainability reports in these regions.

Figure~\ref{fig:geographic_coverage} summarizes the geographic distribution of
building-year records in the dataset. Australia accounts for 72.1\% of records,
driven by the availability of detailed public building and energy datasets,
while Europe contributes 27.9\%, largely sourced from organization-level
disclosures linked to commercial real estate portfolios.


\subsection{Building-Level Availability of Scope 3 Categories}

\begin{figure}[t]
\centering
\includegraphics[width=0.9\linewidth]{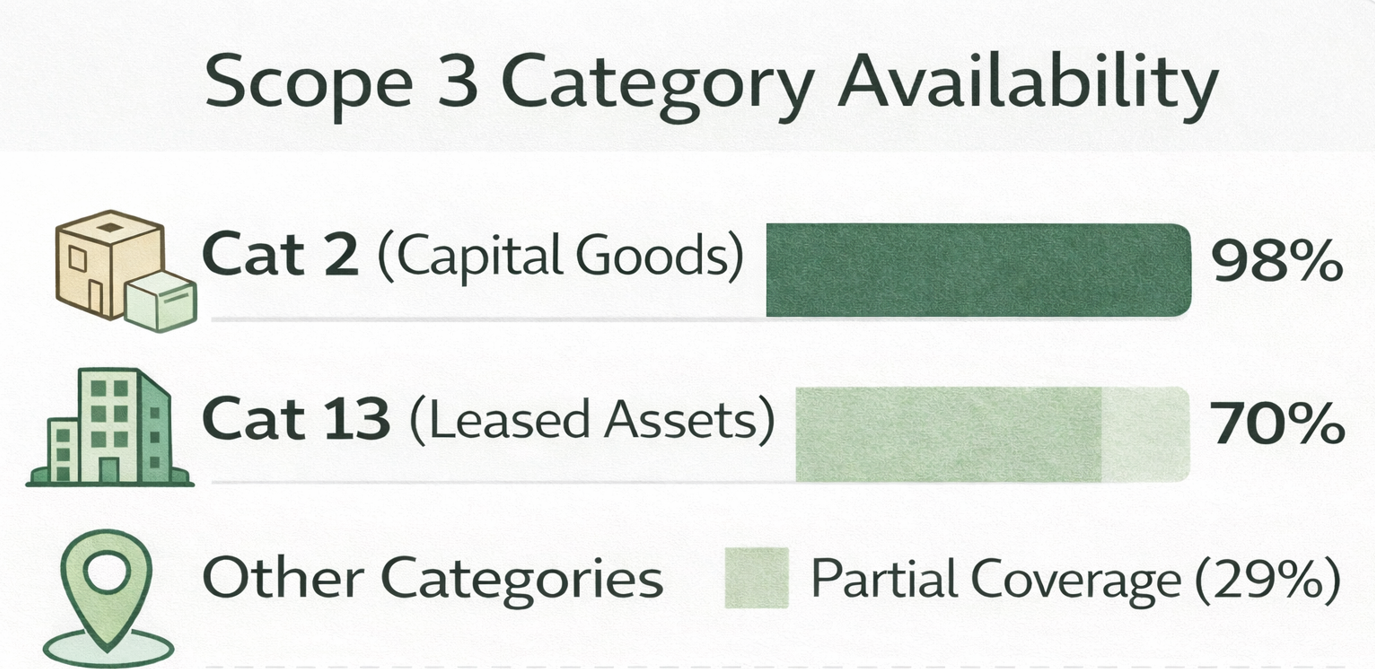}
\caption{
Building-level availability of major Scope~3 emission categories in Scope3Trace.
}
\label{fig:scope3_category_availability}
\end{figure}

Figure~\ref{fig:scope3_category_availability} summarizes the availability of major Scope~3 categories at the building level in Scope3Trace. Categories such as capital goods (Category~2) and downstream leased assets (Category~13) exhibit relatively higher observability due to their physical linkage to building attributes, while most other categories remain only partially observable. This reflects current data availability rather than methodological limitations.

\subsection{Unit Normalization}

All emissions values are standardized to tonnes CO$_2$ equivalent (tCO$_2$e). 
Energy data are harmonized into common units (kWh, MJ, GJ) using standard 
conversion factors to ensure comparability across sources.

\subsection{Intended Applications}

The dataset supports research in:

\begin{itemize}
    \item Scope~3 emissions estimation and benchmarking;
    \item building-level carbon accounting;
    \item multimodal climate data modeling;
    \item corporate sustainability disclosure analysis.
\end{itemize}

\section{Building-Definable Scope 3 Categories: Rationale for Focusing on Category 2 and 13}
\label{sec:why}

Scope 3 emissions encompass heterogeneous value chain activities whose observability varies substantially across categories. Some categories are directly linked to physical assets, while others primarily reflect organizational procurement, logistics, or service activities. Scope3Trace therefore focuses on Scope 3 categories that can be interpreted from asset-level characteristics.

\subsection{Category 2: Capital Goods}

Category 2 emissions largely arise from embodied carbon in long-lived physical assets such as buildings and infrastructure. These emissions scale with measurable physical characteristics including floor area, building typology, and construction period, for which publicly available embodied carbon factor databases exist (e.g., engineering life-cycle datasets). As a result, Category 2 can be represented at the building level through transparent factor-based approximations that explicitly encode assumptions rather than attempting alternative accounting claims. Such representations support explainability of organizational disclosures by linking aggregated reported values to underlying asset structures.

\subsection{Category 13: Downstream Leased Assets}

Category 13 emissions correspond to operational energy use of leased assets, which in real estate portfolios are directly associated with building-level electricity consumption and tenant activity. When building energy baselines or modeled energy datasets are available, Category 13 emissions can be approximated using region-specific grid emission factors. Because these emissions are operationally tied to identifiable physical assets, Category 13 provides a structurally observable bridge between organizational Scope 3 disclosures and asset-level activity signals.

\subsubsection{Why other Scope 3 Categories are not estimated}

Most remaining Scope 3 categories primarily reflect organizational activities rather than asset characteristics. Purchased goods and services (Category 1), upstream energy supply chain emissions (Category 3), waste treatment (Category 5), and travel-related categories depend on procurement practices, supply-chain relationships, or human mobility patterns that are typically not observable from building-level physical datasets. Estimating these categories from asset attributes alone would require strong unverifiable assumptions and risk conflating explanatory modeling with formal emissions accounting. Scope3Trace therefore does not derive building-level estimates for these categories; values for Cat~3 or Cat~5 that appear in our dataset are extracted verbatim from organizational disclosures rather than estimated from building features.

\section{Derivation formulas and units}
\label{sec:formulas}

This appendix summarizes the deterministic derivation formulas used to construct asset-level explanatory signals in Scope3Trace. These derivations are intended to support explainability, coverage diagnostics, and methodological transparency. All derived values are explicitly labeled as modeled or estimated in the dataset.

\subsection{Scope 2 Operational Emissions}

When building-level electricity consumption data are available but Scope 2 emissions are not reported, emissions are approximated by $\text{Scope2}_{tCO_2e} = \text{Electricity}_{kWh} \times EF_{grid}$, where $\text{Electricity}_{kWh}$ denotes annual electricity consumption attributed to the building (in kWh) and $EF_{grid}$ denotes the electricity grid emission factor (in tonnes CO$_2$e per kWh), selected according to geographic region and reporting year.

\subsection{Scope 3 Category 2}

Scope 3 category 2 emissions at the building level are approximated using floor-area-based embodied carbon intensity factor: $\text{Cat2}_{tCO_2e} = \frac{\text{Floor Area}_{m^2} \times EF_{\text{embodied}}}{1000}$, where $\text{Floor Area}_{m^2}$ denotes gross building floor area in square meters, $EF_{\text{embodied}}$ denotes embodied carbon intensity, selected according to building typology where available. 

\subsection{Scope 3 Category 13}

Category 13 emissions correspond to operational emissions from leased assets and are approximated using tenant-attribute electricity consumptions: $\text{Cat13}_{tCO_2e} = \text{Tenant Electricity}_{kWh} \times EF_{\text{grid}}$, where $\text{Tenant Electricity}_{kWh}$ is estimated from total building electricity consumption scaled by the commercial buildings proportion, $EF_{\text{grid}}$ is the same region and year specific grid emission factor used for Scope 2 estimation.

\subsection{Provenance Labels and Interpretation}

All derived values in Scope3Trace are accompanied by structured provenance metadata, including:

\begin{itemize}
    \item reported: directly extracted from disclosures,
    \item modeled: derived from public baseline datasets,
    \item estimated: calculated using emission factors or proxy assumption
\end{itemize}

These labels allow users to distinguish factual disclosures from explanatory approximations and support uncertainty-aware downstream analysis. Importantly, derived values are not intended to replace corporate greenhouse gas accounting results but to make underlying assumptions, asset coverage, and potential evidence gaps transparent.

\section{Computational Overhead Analysis}
\label{app:overhead}

\begin{table}[t]
\centering
\small
\setlength{\tabcolsep}{3pt}
\caption{Per-document overhead of each Scope3Trace agent, averaged over 10 representative reports. Cost is computed using public per-token pricing.}
\label{tab:overhead_per_agent}
\begin{tabular}{lrrrrr}
\toprule
Agent & Time (s) & In tok. & Out tok. & GPT-4o & GPT-4o-mini \\
\midrule
Search & 6.7 & 22{,}886 & 151 & \$0.059 & \$0.004 \\
Scope & 12.7 & 82{,}685 & 1{,}015 & \$0.217 & \$0.013 \\
Verify & 5.3 & 42{,}461 & 303 & \$0.109 & \$0.007 \\
\midrule
Total & 24.7 & 148{,}032 & 1{,}469 & \$0.385 & \$0.024 \\
\bottomrule
\end{tabular}
\end{table}

\begin{table}[t]
\centering
\small
\setlength{\tabcolsep}{3pt}

\begin{tabular}{lrrcc}
\toprule
Method & In tok. & Cost & Cat F1 & EM \\
\midrule
Rule-only & --- & \$0 & 0.51 & 0.45 \\
GPT-4o & 137{,}723 & \$0.351 & 0.88 & 0.82 \\
Gemini-2.0 & 137{,}723 & \$0.014 & 0.87 & 0.81 \\
DeepSeek-V3 & 137{,}723 & \$0.038 & 0.86 & 0.80 \\
CrewAI-style & 165{,}884 & \$0.432 & 0.75 & 0.54 \\
MetaGPT-style & 172{,}193 & \$0.451 & 0.77 & 0.59 \\
\midrule
Scope3Trace (GPT-4o) & 148{,}032 & \$0.385 & \textbf{0.99} & \textbf{0.96} \\
Scope3Trace (GPT-4o-mini) & 147{,}053 & \$0.024 & 0.92 & 0.90 \\
\bottomrule
\end{tabular}

\caption{Per-document overhead and accuracy vs.\ single-call and multi-agent LLM baselines (averaged over the same 10 reports). Generic multi-agent frameworks incur substantially higher coordination overhead, while Scope3Trace achieves the highest extraction accuracy with comparatively modest additional cost.}
\label{tab:overhead_vs_baseline}

\end{table}

We profile Scope3Trace's per-document overhead on 10 representative reports. Table~\ref{tab:overhead_per_agent} reports wall time, token usage, and API cost broken down per agent using both GPT-4o and the cheaper GPT-4o-mini (with the same pipeline structure). Table~\ref{tab:overhead_vs_baseline} compares total per-document overhead against the single-call LLM baselines from Table~\ref{tab:extraction_results}.

Scope3Trace's overhead is comparable to single-call LLM baselines while achieving substantially higher accuracy. Switching to GPT-4o-mini reduces cost by $\sim$16$\times$ with only a moderate accuracy drop, indicating that the rule-bounded pipeline design also enables cost-efficient deployment.

\section{LLM Backbone Robustness}
\label{app:backbone}

Scope3Trace delegates all deterministic logic—numeric parsing, unit
normalization, category mapping, and evidence checking—to rule-based
components, restricting the shared LLM to bounded semantic tasks. To verify
that overall accuracy does not depend on a specific model family, we hold the
pipeline, prompts, and gold evaluation set fixed and swap only the shared LLM
backbone across all three agents. Table~\ref{tab:backbone_robustness} reports
the results over three runs, with per-document cost taken from
Appendix~\ref{app:overhead}.

Our decision to treat the backbone as the only variable rests on the observation that post-training, rather than architecture alone, largely determines the behavior of a deployed LLM. Recent studies analyze when and why supervised fine-tuning (SFT) fails to fully learn from training data \citep{xue2026supervised}, propose anchored SFT to stabilize post-training \citep{zhu2026anchoredsupervisedfinetuning}, study how parameter importance evolves during fine-tuning to guide selective parameter isolation \citep{lin2026parameter}, and report similar limitations in vision-language reasoning \citep{yu2025reassessing}. Complementary work improves post-training through on-policy distillation \citep{xu2026beyond,xu2026tip,xu2026paced}, studies spurious signal amplification in test-time reinforcement learning \citep{yu2026understanding}, and develops reward models and preference selection for alignment \citep{xue2026reason,lai2026unirrm,lai2026aligndiff}. Taken together, this literature implies that conclusions about LLM-based extraction should not depend on the specific post-trained model deployed, which is precisely the question the experiments below address.

Performance remains within a narrow band across all four backbones, supporting
our design rationale: once deterministic logic is offloaded to rules, the
remaining semantic tasks are robust to LLM choice. Notably, the GPT-4o-mini
backbone retains strong accuracy at roughly $16\times$ lower cost, indicating
that the rule-bounded design also enables cost-efficient deployment.

\paragraph{Heterogeneous per-agent configurations.}
The risk of correlated errors from a shared backbone is structurally
mitigated: final numeric verification is performed by deterministic
rules matching each value against the source OCR text, while the Verify
Agent's LLM modules handle semantic validation only and can flag and
return a failing value but never correct it, so no channel exists for
one LLM to approve another's numeric output. Consistent with this,
Table~\ref{tab:hetero_backbones} shows four heterogeneous configurations
staying within 0.03 EM of the homogeneous setting; in particular,
swapping only the Verify Agent to a different model family (third row)
leaves EM unchanged at 0.96.

\begin{table}[t]
\centering
\footnotesize
\setlength{\tabcolsep}{4pt}

\resizebox{\columnwidth}{!}{%
\begin{tabular}{lccc}
\toprule
Search / Scope / Verify & Scope F1 & Cat F1 & EM \\
\midrule
GPT-4o / Gem-2.0 / DS-V3
    & 0.98$\pm$0.01
    & 0.97$\pm$0.01
    & 0.96$\pm$0.01 \\

DS-V3 / GPT-4o / Gem-2.0
    & 0.96$\pm$0.02
    & 0.94$\pm$0.02
    & 0.93$\pm$0.02 \\

GPT-4o / GPT-4o / Gem-2.0
    & 0.99$\pm$0.004
    & 0.98$\pm$0.006
    & 0.96$\pm$0.01 \\

Gem-2.5-F / DS-Chat / GPT-4o
    & 0.98$\pm$0.01
    & 0.97$\pm$0.01
    & 0.96$\pm$0.01 \\
\midrule

Homogeneous GPT-4o (Table~\ref{tab:backbone_robustness})
    & 0.99$\pm$0.004
    & 0.97$\pm$0.008
    & 0.96$\pm$0.01 \\
\bottomrule
\end{tabular}%
}

\caption{Heterogeneous per-agent backbones; pipeline, prompts, and gold
set fixed. Gem $=$ Gemini, DS $=$ DeepSeek, Gem-2.5-F $=$
Gemini-2.5-Flash.}
\label{tab:hetero_backbones}
\end{table}

\begin{table}[t]
\centering
\small
\resizebox{\columnwidth}{!}{%
\begin{tabular}{lcccc}
\toprule
LLM Backbone & Scope F1 & Cat F1 & EM & Cost/doc \\
\midrule

GPT-4o 
& 0.99$\pm$0.004 
& 0.97$\pm$0.008 
& 0.96$\pm$0.01 
& \$0.385$\pm$0.006 \\

GPT-4o-mini 
& 0.97$\pm$0.008 
& 0.94$\pm$0.015 
& 0.91$\pm$0.02 
& \$0.024$\pm$0.002 \\

Gemini-2.0 
& 0.98$\pm$0.006 
& 0.95$\pm$0.012 
& 0.93$\pm$0.02 
& \$0.015$\pm$0.001 \\

DeepSeek-V3 
& 0.97$\pm$0.010 
& 0.93$\pm$0.018 
& 0.90$\pm$0.02 
& \$0.041$\pm$0.003 \\
\bottomrule
\end{tabular}}
\caption{Scope3Trace performance with different shared LLM backbones; pipeline, prompts, and gold set held fixed. Mean $\pm$ std over 3 runs. Cost/doc from Appendix~\ref{app:overhead}.}
\label{tab:backbone_robustness}
\end{table}

\section{Setup for RAG and Multi-Agent Comparisons}
\label{app:autogen}

Quantitative results for the retrieval-augmented and multi-agent baselines reported in Table~\ref{tab:extraction_results} are produced under matched conditions described here.

\paragraph{\citet{Beck2025}-style baseline.}
We re-implement the retrieval-augmented (RAG) pipeline of
\citet{Beck2025}: pages are chunked page-wise into a vector store,
retrieved by cosine similarity against their fixed query
(text-embedding-ada-002; unspecified in the original), and the top-3
pages plus each's preceding and following page form the context. Two
deliberate deviations ensure a matched comparison: GPT-4o
(temperature $=0$) replaces GPT-4, and our Scope Agent's prompt replaces
their question template, isolating pipeline architecture from
backbone and prompt effects. There is no table reconstruction or
rule-bounded numerical verification.

\paragraph{Multi-agent baselines.} For \textbf{AutoGen}~\citep{wu2023autogen},
\textbf{CrewAI}~\citep{crewai}, and \textbf{MetaGPT}~\citep{metagpt}, we use the same 3-agent role decomposition (Search / Scope / Verify), the same underlying LLM (GPT-4o, temperature=0), the same prompts, and the same 200-document gold evaluation set as Scope3Trace.

\paragraph{Validation against official libraries.}
AutoGen, CrewAI, and MetaGPT are general-purpose orchestration frameworks
with no official pipeline for Scope~1--3 extraction, so task-specific
roles and prompts must be authored regardless of implementation. To verify
that our re-implementations do not understate them, we re-ran the official
AutoGen (v0.7.5) and CrewAI (v1.15.2) libraries under the identical
configuration above; results match within $\pm0.02$~EM
(Table~\ref{tab:official_libs}), confirming the gap is architectural
rather than implementational. MetaGPT's official library hard-codes a
software-engineering SOP that cannot be reconfigured for this task without
rewriting its orchestration core, so we retain the ``-style''
re-implementation for it. All baseline code, configurations, and framework
versions are released with our repository.

\begin{table}[t]
\centering
\small
\begin{tabular}{llc}
\toprule
Baseline & Backbone & EM \\
\midrule
AutoGen-style (ours)    & GPT-4o & 0.48 \\
AutoGen official v0.7.5 & GPT-4o & 0.50 \\
\midrule
CrewAI-style (ours)     & GPT-4o & 0.54 \\
CrewAI official v1.15.2 & GPT-4o & 0.53 \\
\bottomrule
\end{tabular}
\caption{Re-implemented baselines vs.\ official libraries under identical
configurations (cf.\ Table~\ref{tab:extraction_results}).}
\label{tab:official_libs}
\end{table}

\paragraph{Diagnostic for multi-agent underperformance.} Even with identical role decomposition, the same LLM, and the same prompts, generic conversational orchestration substantially underperforms Scope3Trace (Table~\ref{tab:extraction_results}). Inspection of failed records shows that noise introduced during free-form agent interaction repeatedly overwrites originally correct values in the verification round. This indicates that Scope3Trace's advantage stems from \emph{combining role-decomposed agents with rule-bounded numerical resolution and evidence-grounding constraints}, rather than from the multi-agent decomposition alone.

\paragraph{Single-call CoT baseline.}
We also test whether a well-designed chain-of-thought prompt in a single
call matches the pipeline: the prompt walks through page localization,
table interpretation, and evidence-anchored extraction step by step
(same setup as Table~\ref{tab:extraction_results}). As shown in
Table~\ref{tab:cot_baseline}, CoT improves semantic interpretation
(category F1 $+0.06$) but \emph{degrades} exact match (0.72
$\rightarrow$ 0.67): longer chains add generation steps in which numeric
transcription drifts. Since the matched multi-agent baselines also
underperform the single call, the advantage stems from rule-bounded
components rather than task decomposition in either direction.

\begin{table}[t]
\centering
\footnotesize
\setlength{\tabcolsep}{3pt}
\begin{tabular}{lccc}
\toprule
Method & Scope F1 & Cat F1 & EM \\
\midrule
Single-call GPT-4o
    & 0.87$\pm$0.02
    & 0.79$\pm$0.03
    & 0.72$\pm$0.03 \\
\quad + CoT
    & 0.89$\pm$0.02
    & 0.85$\pm$0.03
    & 0.67$\pm$0.04 \\
Scope3Trace
    & 0.99$\pm$0.004
    & 0.97$\pm$0.008
    & 0.96$\pm$0.01 \\
\bottomrule
\end{tabular}

\caption{Single-call chain-of-thought baseline vs.\ Scope3Trace on the
gold set. CoT improves category F1 but degrades exact match.}
\label{tab:cot_baseline}
\end{table}

\section{Error Analysis}
\label{app:error_analysis}

\paragraph{Extraction errors.}
Scope3Trace errs on 8 of the 200 gold documents: (a) complex
table-structure misalignment, where merged or spanning cells and
multi-level headers cause row labels, years, and values to be
mis-associated (5 cases); and (b) OCR corruption in low-quality scans,
including digit--letter substitutions (``0/O'', ``1/I'') and malformed
separators or units (3 cases), a failure mode that character-level error correction methods \citep{lin2026cec} are designed to mitigate. No error stems from the rule-bounded
components (category mapping, unit normalization, evidence
verification); all trace to input quality, consistent with the
difficulty stratification in Appendix~\ref{app:goldset_validation}.

\paragraph{Entity linking.}
Of 212 unique responsible entities, 204 (96.2\%) link to
organization-level disclosures. The 8 unmatched cases are 5 private
owners with no public disclosures and 3 SPV/trustee entities whose
legal names diverge substantially from parent brands (e.g., ``Cbus
Property 1 William Street Pty Ltd'' vs.\ ``Cbus Property''), exposing
the limits of lexical and embedding similarity and motivating the
corporate-registry graphs discussed in the Limitations. Text matching and question-answering methods improve the localization and interpretation of relevant content in heterogeneous disclosures \citep{xue2024question,xue2023dual,xue2025structcoh}, but legal-name divergence across registries and disclosures remains an open challenge.

\section{Annotation Guidelines For Gold Dataset Construction}
\label{sec:annotation_instructions}

This appendix documents the evaluation setup and the annotation protocol used to construct the gold evaluation dataset for Scope 1/2/3 emissions extraction and is provided for completeness, the main paper is fully understandable without this section.

\subsection{Extraction evaluation setup}

\paragraph{Gold Set Statistics.} The gold evaluation set contains \textbf{200 reports} spanning \textbf{180 organizations}, \textbf{11 sectors}, and \textbf{4 reporting years}, yielding \textbf{686 Scope value instances} and \textbf{1{,}266 category-level annotations}. Annotations were produced by \textbf{2 annotators} following the protocol described in the subsections below (\S\ref{subsec:annotation_schema}--\S\ref{subsec:annotation_qc}), with \textbf{83\% of documents double-annotated}. Annotators include co-authors; annotation reliability is supported by: (i) all extracted values are numerical figures verifiable against the source PDFs; (ii) strict exclusion criteria (\S\ref{subsec:annotation_exclusion}) minimize subjective judgment calls; and (iii) every page-level evidence link is third-party verifiable against the public source PDFs. On the double-annotated subset, inter-annotator agreement reaches Cohen's $\kappa = 0.89$, indicating almost-perfect agreement under the Landis--Koch interpretation.

The evaluation covers Scope~1, Scope~2, Scope~3 emissions, as well as category-level extraction and exact match (EM) accuracy. For each document, the system processes the full PDF through the Scope3Trace pipeline, including page localization, table reconstruction, structured extraction, and evidence grounding. Ablation variants are constructed by removing individual components (e.g., page localization, table reconstruction, or grounding) while keeping all other modules unchanged.

We compare our method against a rule-based extractor and two LLM-only regimes: a \emph{chunked LLM} baseline that runs per-page calls over OCR text, and a \emph{single-call LLM} baseline that uploads the native PDF and makes one extraction call. Both LLM regimes are evaluated across three model families (GPT-4o, Gemini-2.0, and DeepSeek-V3). For LLM-based methods, we use a fixed prompt template across all models to ensure a fair comparison.

All experiments are conducted with deterministic settings where possible. For LLM-based extraction, we use temperature=0 to reduce randomness. Results are reported as mean $\pm$ standard deviation over three runs.

Evaluation metrics include F1 scores for Scope-level and category-level extraction, as well as exact match (EM) accuracy. Metrics are computed by comparing extracted values against gold annotations at the building level.

\paragraph{Annotation timeline and independence from development.} Annotation guidelines were finalized first, derived from the GHG Protocol Corporate Standard, before any document was labeled; the exclusion criteria (Appendix \ref{subsec:annotation_exclusion}) therefore reflect the Protocol's accounting definitions rather than observed system behavior. The 200-report gold set was then fully annotated and frozen, with no value modified thereafter. All pipeline rules, prompts, and thresholds were developed exclusively on a separate 20-PDF development set with no overlap with the gold set, which was processed only once, for the final evaluation reported in Table \ref{tab:extraction_results}.

\subsection{Annotation Schema}
\label{subsec:annotation_schema}

Annotations follow one of two structured JSON schemas depending on the reporting level.

\paragraph{Organization-level Schema.} Each document yields Scope 1/2/3 totals, a list of Scope 3 category items, evidence text.

\paragraph{Building-level.} Each document corresponds to a single building and includes Scope 1/2/3 totals, Scope 3 categories when disclosed, and normalized address identifiers for downstream multimodal alignment.

\subsection{Scope Extraction Rules}

\paragraph{Scope Totals.}
\begin{itemize}
    \item Scope~1, Scope~2, and Scope~3 totals are extracted only when explicitly
    reported in absolute units (e.g., tCO$_2$e).
    \item Both location-based and market-based Scope~2 values are recorded when available.
    \item Totals are not inferred from percentages, charts, or cross-page aggregation.
\end{itemize}

\paragraph{Scope~3 Category Mapping.}
Each Scope~3 line item is mapped to one of the 15 GHG Protocol categories.
Annotators do not infer undisclosed categories; missing categories are recorded as null.

\subsection{Exclusion Criteria}
\label{subsec:annotation_exclusion}

The following items are explicitly excluded from annotation:
\begin{itemize}
    \item Carbon offsets, credits, RECs, LGCs, or certificates
    \item Intensity metrics (e.g., kgCO$_2$e/m$^2$)
    \item Reduction targets, baselines, or percentage-only disclosures
    \item Combined totals (e.g., Scope~1+2)
    \item Items explicitly labeled as Scope~1 or Scope~2
\end{itemize}

\subsection{Evidence Grounding Requirements}

For each extracted value, annotators must provide:
\begin{itemize}
    \item The exact page number where the value appears
    \item A verbatim text excerpt (100--400 characters) containing the numeric value
    \item For tables, the full row containing the value and at least one adjacent row
\end{itemize}

Evidence text must originate solely from the specified page and must not be
paraphrased or summarized.

\subsection{Year Alignment}
\label{subsec:annotation_year}

\begin{itemize}
    \item Each extracted value is explicitly associated with its reporting year.
    \item For multi-year tables, values for all reported years are extracted separately.
    \item Historical values are preserved rather than overwritten by the latest year.
\end{itemize}

\subsection{Quality Control}
\label{subsec:annotation_qc}

To ensure annotation reliability:
\begin{itemize}
    \item All annotations follow a fixed schema with mandatory fields
    \item Numeric values are validated for unit consistency and formatting
    \item A subset of documents is double-annotated, with disagreements resolved by consensus
\end{itemize}

\subsection{Independent Validation of Gold-Set Reliability}
\label{app:goldset_validation}
 
To rule out annotator bias, easy-document inflation, and development
leakage, we conducted three independent checks after the pipeline and gold
set were frozen. The two external annotators involved in the following validation experiments were university student volunteers recruited through the authors’ academic network. They received no monetary compensation, and their participation was voluntary and unrelated to any course or employment requirement.
 
\paragraph{(a) External re-annotation.}
Two annotators outside the author list independently re-labeled a random
30-report subset of the gold set under the same protocol
(\S\ref{subsec:annotation_schema}--\S\ref{subsec:annotation_year}). Agreement with the
original labels is $\kappa = 0.88$, matching the original inter-annotator
agreement ($\kappa = 0.89$) and indicating no author-specific labeling
bias.
 
\paragraph{(b) Difficulty stratification.}
Table~\ref{tab:difficulty_strata} reports performance by document
difficulty. Errors concentrate in the objectively hard strata (scanned
PDFs, complex multi-page tables), a pattern inconsistent with memorization
or leakage, which would inflate scores uniformly across strata. Even the
hardest stratum's EM (0.85) exceeds the best baseline's aggregate EM
(0.72).
 
\begin{table}[t]
\centering
\small
\begin{tabular}{lccc}
\toprule
Subset & $N$ & Scope F1 & EM \\
\midrule
Native PDF                     & 180 & 0.99 & 0.972 \\
Scanned (OCR)                  &  20 & 0.94 & 0.850 \\
\midrule
Multi-page/multi-year tables   &  55 & 0.97 & 0.909 \\
Simple tables                  & 145 & 0.99 & 0.979 \\
\bottomrule
\end{tabular}
\caption{Gold-set extraction performance stratified by document
difficulty. The two stratifications are orthogonal partitions of the same
200 reports: by PDF type (top) and by table complexity (bottom).}
\label{tab:difficulty_strata}
\end{table}
 
\paragraph{(c) Out-of-distribution evaluation.}
We collected 30 additional reports from companies entirely absent from the
dataset, verified against both the gold and development sets. These were
annotated by the two external annotators, and the frozen pipeline was
applied without modification. Scope3Trace achieves Scope F1 $= 0.98$ and
EM $= 0.93$, within 0.01/0.03 of the gold-set results and above every
baseline's in-distribution EM (Table~\ref{tab:extraction_results}), confirming
that performance generalizes beyond the evaluation distribution.

\subsection{Ethical Considerations}

All annotated documents are publicly available corporate disclosures.
The annotation process involves no personal or sensitive data and poses minimal risk.
Annotations contain only aggregated emissions values and organizational metadata and
are used exclusively for research and evaluation purposes.

\subsection{Example Annotation}

\begin{figure}[t]
\centering
\scriptsize
\begin{verbatim}
{
  "org_id": "ORG_EXAMPLE",
  "org_name_standardized": "Example REIT",
  "report_year": 2023,
  "reporting_boundary": "Operational Control (Assumed)",
  "assurance_level": "Unknown",
  "source_document": "example-sustainability-report-2023.pdf",
  "schema_version": "v2.1",

  "scope1_tco2e_value": 221.0,
  "scope1_provenance": "reported",

  "scope2_location_tco2e_value": 0.0,
  "scope2_market_tco2e_value": null,
  "scope2_provenance": "reported",

  "scope3_summary": {
    "total_tco2e_value": 221.0,
    "provenance": "reported",
    "breakdown": {
      "cat2_capital_goods_tco2e": 17.0
    }
  },

  "evidence_pages": [3, 4, 6]
}
\end{verbatim}
\caption{
Illustrative example of a gold annotation record.
Each extracted value is linked to a reporting year and a source page.
Evidence text is stored verbatim in the dataset but omitted here for brevity.
}
\label{fig:annotation_example}
\end{figure}

Figure~\ref{fig:annotation_example} shows an example building-level annotation illustrating Scope totals, a Scope 3 category item, and page-level evidence.

\section{Factor sources and parameter provenance}
\label{sec:FactorSources}

Scope3Trace incorporates publicly documented emission factors and activity proxies to support transparent interpretation of building-level Scope 3 components. These factors are not intended to provide alternative accounting estimates but to expose underlying assumptions commonly used in industry Scope 3 reporting practices. 

Three factor families are used. Grid factors ($EF_{grid}$) follow a
state $\times$ reporting-year lookup from the Australian National
Greenhouse Accounts (NGA) Factors, 2019 to 2025 editions
\citep{nga2019,nga2025}, defaulting conservatively to the highest-factor
state when state information is missing. Embodied factors
($EF_{embodied}$, Category~2) use a deterministic lookup of fixed
coefficients over 11 building typologies, calibrated to published
whole-building ranges. Commercial shares ($p_{com}$, Category~13) come
from the City of Melbourne property-level energy dataset
\citep{com_energy}, matched to the model year. Complete lookup tables
with values, units, sources, versions, and access dates are released
with v1.0.

All factor applications follow deterministic unit normalization and conversion procedures to ensure reproducibility. Units are standardized to $CO_2e$, and activity measures such as electricity consumption and floor area are converted using documented conventions. Detailed derivation formulas and unit conventions are provided in Appendix~\ref{sec:formulas}

\section{Organization-level Detailed Statistics}
\label{sec:org_appendix}

This appendix provides supplementary information on the organization-level component of Scope3Trace, and this section documents the provenance sectoral distribution, and disclosure characteristics of organization-level Scope3 data to support transparency and reproducibility.

Records are collected from publicly available sustainability, climate, and greenhouse gas inventory reports. These disclosures typically present aggregated Scope 1-3 emissions together with category-level breakdowns aligned with the GHG Protocol. Because reporting practices vary across industries and jurisdictions, this component of the dataset reflects heterogeneous disclosure granularity, assurance levels, and reporting boundaries. 

\begin{figure}[t]
    \centering
        \includegraphics[width=0.95\linewidth]{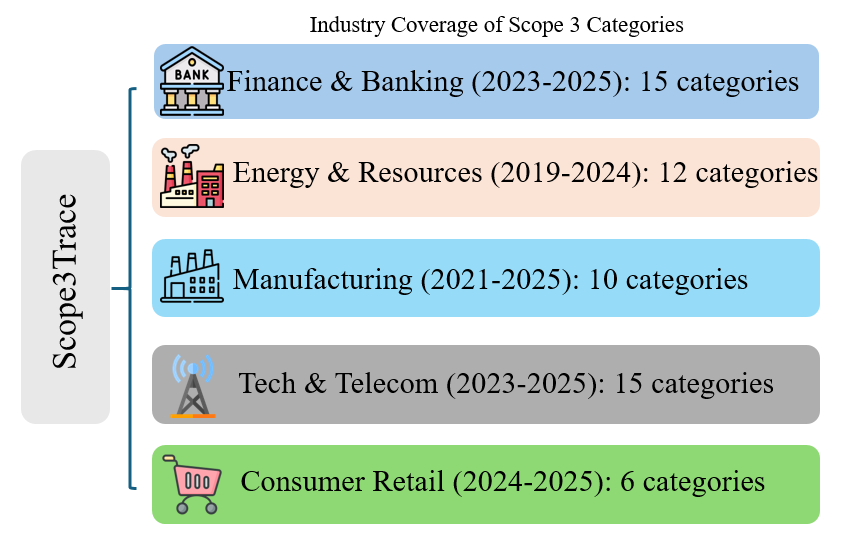}
    \caption{
    Organization-level dataset profile in Scope3Trace. 
The dataset spans multiple sectors, jurisdictions, and reporting years, 
reflecting heterogeneous real-world Scope~3 disclosure practices.
    }
    \label{fig:org_dataset_profile}
\end{figure}

Table~\ref{tab:org_dataset_profile} summarizes the sectoral coverage, reporting years, and category disclosure characteristics represented in the dataset, visualized in Figure~\ref{fig:org_dataset_profile}. The included organizations span multiple sectors, and cover several geographic regions. Reporting years range primarily from 2019 to 2025, reflecting the availability of recent sustainability disclosures.

Table~\ref{tab:org_report_characteristics} provides illustrative examples highlight variation in category completeness, third-party assurance, and regional reporting conventions. Such heterogeneity is intentionally preserved in Scope3Trace, as it reflects real-world reporting conditions and enables analysis of disclosures practices.

Organization-level disclosures in Scope3Trace are treated as reported control totals. They provide contextual reference values that support explainability, boundary diagnostics, and transparency analysis. 

\begin{table*}[t]
\centering
\small
\setlength{\tabcolsep}{7pt}
\caption{Organization-level dataset profile in Scope3Trace. 
The dataset spans multiple sectors, jurisdictions, and reporting years, 
reflecting heterogeneous real-world Scope~3 disclosure practices.}
\label{tab:org_dataset_profile}
\begin{tabular}{
p{3.6cm}
p{2.1cm}
p{2.3cm}
p{6.0cm}
}
\toprule
\textbf{Sector} & 
\textbf{Organizations} & 
\textbf{Reporting Years} & 
\textbf{Scope~3 Categories Disclosed} \\
\midrule
Finance and Banking 
& Multiple 
& 2023--2025 
& Mostly categories 1--15 \\

Energy and Resources 
& Multiple 
& 2019--2024 
& Partial to full category coverage \\

Manufacturing and Industry 
& Multiple 
& 2021--2025 
& Typically categories 1--10 or full disclosure \\

Technology and Telecommunications 
& Multiple 
& 2023--2025 
& Mostly categories 1--15 \\

Consumer Goods and Retail 
& Multiple 
& 2024--2025 
& Category-level disclosures common \\
\bottomrule
\end{tabular}
\end{table*}

\begin{table*}[t]
\centering
\setlength{\tabcolsep}{3.5pt}
\caption{Example characteristics of organization-level sustainability disclosures included in Scope3Trace. Reports vary in category granularity, assurance level, and geographic origin.}
\label{tab:org_report_characteristics}
\scalebox{0.9}{
\begin{tabularx}{\textwidth}{l c >{\RaggedRight\arraybackslash}X c c}
\toprule
Organization & Reporting Year & Categories Disclosed & Assurance Level & Region \\
\midrule
BASF & 2023 & 1--15 & Third-party assurance & Europe \\
Singtel Group & 2025 & 1--15 & Limited assurance & Asia-Pacific \\
Macquarie Group & 2025 & Category breakdown provided & Limited assurance & Australia \\
Boston Scientific & 2024 & Category-level disclosure & Third-party assurance & North America \\
Trafigura & 2025 & Multi-category disclosure & Not specified & Global \\
\bottomrule
\end{tabularx}
}
\end{table*}

\subsection{Restated Disclosures}
\label{app:restatements}
Corporate reports occasionally restate prior-year figures. Scope3Trace
preserves and traces rather than reconciles: original and restated
observations coexist, each carrying its own source document, page, and
verbatim evidence. In v1.0, 2.3\% of organization-year pairs have
multi-vintage observations; all 36 such pairs show discrepancies, with a
median relative difference of 19.1\%, predominantly in Scope~3. The
schema marks these via the \texttt{restated} flag and
\texttt{superseded\_by} link (Appendix~\ref{appendix:dataset_schema}),
allowing users to select either the latest or the as-first-reported
vintage without losing provenance.

\section{Value Provenance and Data Quality Diagnostics}
\label{sec:valueProvenance}

This appendix provides detailed statistics on value provenance, metadata completeness, 
missingness patterns, and confidence annotations in the building-level component of 
Scope3Trace. These diagnostics complement the summary presented in the main paper and 
are intended to support transparent interpretation of the dataset rather than to serve 
as accuracy benchmarks.

The verification step of our pipeline casts LLMs as judges that cross-check each extracted value against its source evidence, and the confidence annotations reported in this appendix build on this design. Recent studies of LLM-as-a-judge evaluation have examined systematic bias in judge assessment \citep{lai2026biasscopeautomateddetectionbias} and improved alignment between judge and human judgments through internal representations \citep{lai2025surfaceenhancingllmasajudgealignment}; these findings motivate the evidence-based judging criteria used in our verification and confidence labeling.

\subsection{Provenance Distribution}

\begin{figure}[t]
    \centering
    \includegraphics[width=\linewidth]{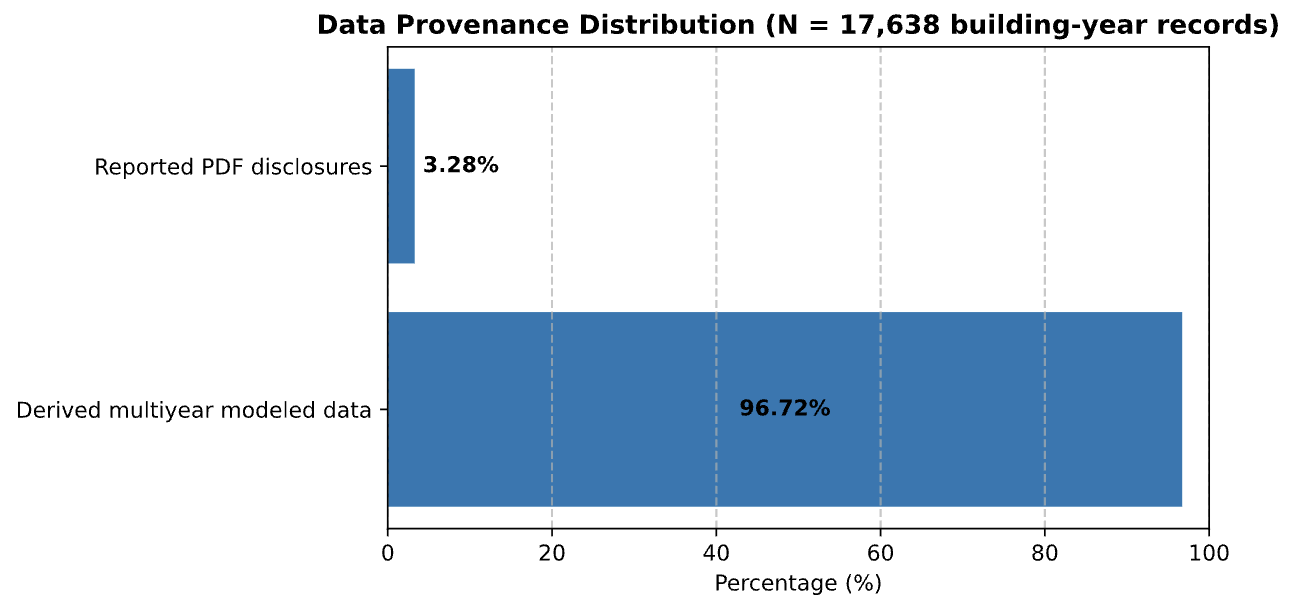}
    \caption{
    Data provenance distribution (non-residential buildings).
    }
    \label{fig:prov_dist}
\end{figure}

\begin{figure}[t]
    \centering
    \includegraphics[width=\linewidth]{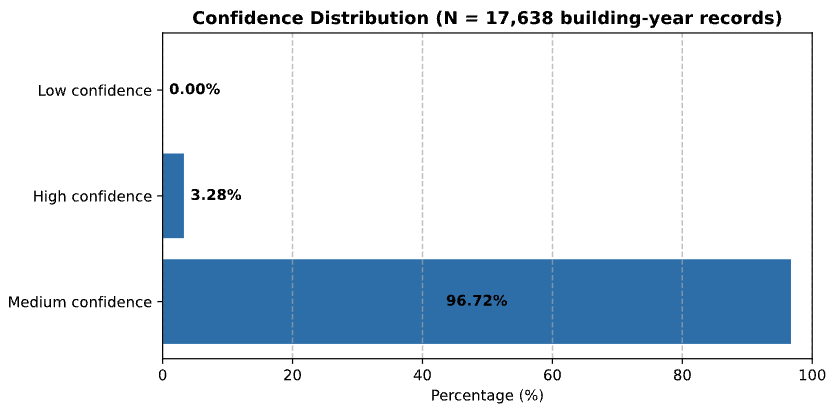}
    \caption{
    Value provenance and confidence summary for non-residential buildings in Scope3Trace.
    }
    \label{fig:value_confidence_summary}
\end{figure}

Figure~\ref{fig:prov_dist} summarizes the distribution of data sources for non-residential 
building records used in Scope~3 analysis. Only a small fraction of building-level 
emissions originates from direct sustainability disclosures, while the majority is 
derived from modeled energy inventories combined with publicly documented emission 
factors. This reflects current industry reporting practices where building-level 
Scope~3 disclosure remains sparse. Figure~\ref{fig:value_confidence_summary} 
presents provenance and confidence side by side, showing that the two 
dimensions coincide: reported values carry high confidence and derived 
values medium confidence.

\subsection{Provenance Metadata Completeness}

Scope3Trace explicitly records provenance metadata including estimation method, 
data source, and quality indicators. Table~\ref{tab:metadata_availability} shows 
field availability rates. Most provenance indicators are nearly complete, although 
method-specific metadata is less consistently available for certain derived Scope~3 
categories due to upstream dataset limitations.

\begin{table}[h]
    \centering
    \setlength{\tabcolsep}{20pt}
    \caption{Availability of provenance metadata fields.}
    \label{tab:metadata_availability}
    \scalebox{0.9}{
        \begin{tabular}{lr}
            \toprule
            Field & Availability \\
            \midrule
            scope2\_quality & 100.0\% \\
            scope2\_method & 76.1\% \\
            scope2\_source & 100.0\% \\
            scope3\_quality & 100.0\% \\
            scope3\_source & 100.0\% \\
            scope3\_cat2\_method & 99.9\% \\
            scope3\_cat13\_method & 70.5\% \\
            \bottomrule
        \end{tabular}
    }
\end{table}

\subsection{Missingness Characteristics}

Table~\ref{tab:missing_fields} reports the most frequently missing numeric variables. 
Higher missingness in tenant electricity and Scope~2 emissions is primarily associated 
with incomplete activity data in upstream government inventories rather than errors 
introduced during dataset construction. Importantly, missing values are retained with 
explicit provenance annotations to support uncertainty-aware analysis.

\begin{table}[t]
\centering
\setlength{\tabcolsep}{4pt}
\caption{Top missing numeric fields.}
\label{tab:missing_fields}
\scalebox{0.9}{
\begin{tabularx}{\columnwidth}{>{\RaggedRight\arraybackslash}X r}
\toprule
Field & Missing Rate \\
\midrule
Scope~3 Cat13 emissions & 29.5\% \\
Scope~2 emissions & 23.8\% \\
Floor area & 0.1\% \\
Scope~3 Cat2 emissions & 0.1\% \\
Electricity energy data & 0.0\% \\
\bottomrule
\end{tabularx}
}
\end{table}

\subsection{Confidence Annotation Characteristics}

Confidence levels in Scope3Trace are designed to indicate provenance reliability 
rather than numerical accuracy. High-confidence values correspond primarily to 
directly reported disclosures, whereas medium-confidence values represent 
deterministic estimates based on activity proxies and public emission factors.

\begin{figure}[t]
    \centering
    \includegraphics[width=\linewidth]{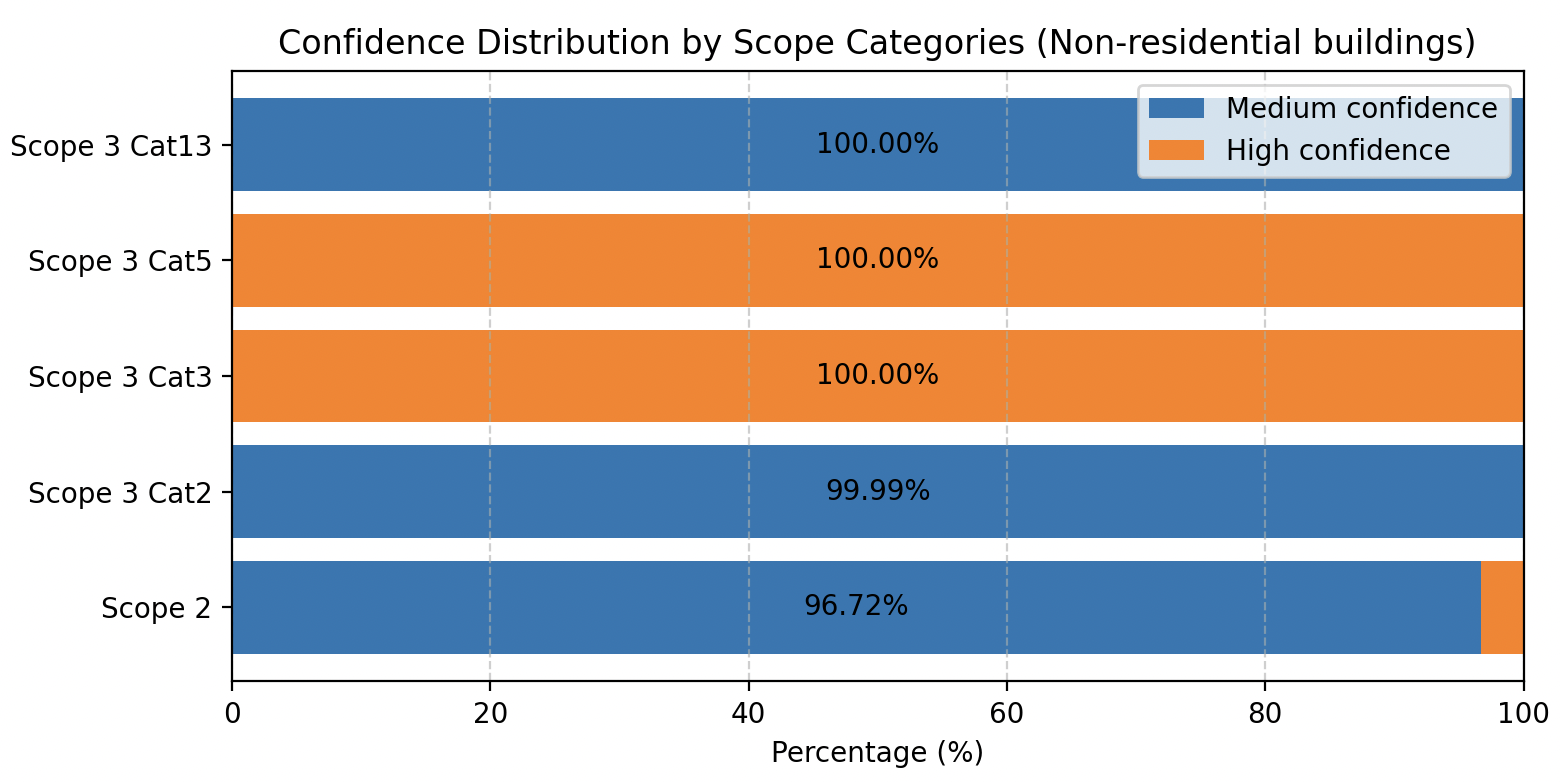}
    \caption{
    Confidence distribution by Scope categories (non-residential).
    }
    \label{fig:confidence_distribution}
\end{figure}

Table~\ref{tab:confidence_distribution} summarizes confidence characteristics 
for key Scope categories in non-residential buildings, visualized in
Figure~\ref{fig:confidence_distribution}.

\begin{table}[t]
\centering
\caption{Confidence distribution by Scope categories (non-residential).}
\label{tab:confidence_distribution}
\resizebox{\columnwidth}{!}{%
\begin{tabular}{lrr}
\toprule
Category & High Confidence & Medium Confidence \\
\midrule
Scope 2 & 3.28\% & 96.72\% \\
Scope 3 Category 2 & 0.01\% & 99.99\% \\
Scope 3 Category 13 & 0\% & 100\% \\
\bottomrule
\end{tabular}%
}
\end{table}

Overall, these diagnostics highlight that Scope3Trace prioritizes transparency of 
data provenance and interpretability of emissions signals over strict completeness. 
The dataset explicitly preserves uncertainty sources, enabling downstream research 
on data coverage, interpretability, and reporting practices.

\section{Physical Plausibility and Consistency Checks}
\label{sec:physical_plausibility}

Physical consistency checks ensure that reported and derived values align with observable activity proxies and publicly documented emission factors. We use a broader dataset containing valid Scope~2 emissions and energy activity data (N = 49{,}322), stratified into commercial (N = 12{,}035) and residential (N = 37{,}287) subsets.

\begin{itemize}[noitemsep, topsep=0pt]
\item Implied emission factors derived from Scope~2 emissions and electricity consumption fall within expected regional grid-factor ranges.
\item Energy use intensity (EUI) checks identify approximately 2\% outliers in both commercial and residential subsets. Most anomalies originate from missing floor-area values in upstream datasets.
\item About 6.1\% of commercial buildings contain missing or invalid floor-area data inherited from government baseline inventories, while residential records show negligible violations.
\end{itemize}

Overall, these diagnostics indicate that Scope3Trace emissions values are physically interpretable and internally consistent while preserving upstream data limitations.

\section{Dataset Summary Tables}
\label{sec:dataset_summary_tables}

This appendix provides a concise statistical overview of the Scope3Trace 
dataset to complement the analyses presented in the main text. These 
summary tables are intended to give readers a quick snapshot of dataset 
scale, temporal and geographic coverage, and the provenance composition 
of emission values. Detailed methodological explanations and diagnostic 
analyses are provided in earlier sections and appendices.

\begin{table}[t]
\centering
\caption{Dataset scale and coverage summary for Scope3Trace.}
\resizebox{\columnwidth}{!}{
\scalebox{0.9}{
\begin{tabular}{lr}
\toprule
Metric & Value \\
\midrule
Unique buildings & $\sim$4,300 \\
Building-year records (with Scope2/3 data) & 17,638 \\
Total building records (all years) & 54,361 \\
Temporal coverage & 2011--2026 \\
Geographic coverage & Australia and Europe \\
Organization-level records & 1,556 organization-year records (v1.0) \\
\bottomrule
\end{tabular}
}
}
\end{table}

\begin{table}[t]
\centering
\caption{Provenance distribution of emission values across major scopes and selected Scope~3 categories (non-residential subset). \emph{Reported}: extracted verbatim from source disclosures; \emph{Estimated/Modeled}: derived from building-level features via Scope3Trace's deterministic pipeline. Cat~2 and Cat~13 are the only categories with a building-level derivation pipeline (Appendix~\ref{sec:why}); Cat~3 and Cat~5 values appear only when directly reported.}
\label{tab:provenance}
\resizebox{\columnwidth}{!}{
\begin{tabular}{lrr}
\toprule
Category & Reported (\%) & Estimated/Modeled (\%) \\
\midrule
Scope 1 & 3.28 & 96.72 \\
Scope 2 & 3.28 & 96.72 \\
Scope 3 (overall) & 3.28 & 96.72 \\
Scope 3 Category 2 & 0.01 & 99.99 \\
Scope 3 Category 3 & 100.0 & 0.0 \\
Scope 3 Category 5 & 100.0 & 0.0 \\
Scope 3 Category 13 & 0.0 & 100.0 \\
\bottomrule
\end{tabular}
}
\end{table}

These statistics reflect the current state of publicly available 
building-level Scope~3 disclosures, where direct reporting remains 
relatively sparse and estimation-based explanatory signals are 
commonly required to support interpretation and coverage analysis. 
Provenance indicators in Scope3Trace explicitly distinguish reported 
values from modeled or proxy-derived signals to enable transparent 
downstream use.

\section{Building-Level Forecasting Benchmark}
\label{app:prediction_benchmark}

\subsection{Task Definition}

To evaluate the practical utility of the constructed dataset in downstream applications, we formulated a building-level Scope 3 emissions prediction task. Given building metadata and satellite imagery, the objective is to predict the total Scope 3 emissions and category-level emissions associated with each building.

Input features include structured building attributes (e.g., floor area, number of levels, and geographic location), as well as satellite imagery capturing the surrounding built environment. The prediction targets are the total Scope 3 emissions and categorized emissions for each building. 

This task enables us to assess whether the collected multimodal data provides useful signals for estimating building-level Scope 3 emissions.

\subsubsection{Prediction Baselines}

To evaluate the effectiveness of our proposed method, we compared it with several baseline models representing different modeling paradigms. These baseline models include traditional tabular regression models, image-based models, and multimodal models that combine both modalities. The aim is to assess the predictive capabilities of building metadata, satellite imagery, and their combinations in forecasting Scope 3 emissions.

\paragraph{Tabular-only Models.} As a baseline for structured metadata, we trained regression models that rely solely on building attribute tables (e.g., floor area, geographic coordinates). We considered a variety of regression models commonly used for structured data, including Ridge regression, Random Forests, and XGBoost.

\paragraph{Image-only Models.} To evaluate the predictive power of visual information alone, we trained the model using satellite images without the tabular metadata. We extracted image representations using a pre-trained convolutional neural network encoder and used the resulting image embeddings as input for emission predictions.

\paragraph{Multimodal.} To investigate whether combining architectural metadata with satellite imagery can improve predictive performance, we developed a multimodal model that fuses tabular features with image embeddings. Specifically, we concatenate the extracted image features with the tabular features and feed them into a downstream regression model.

\subsubsection{Experimental Setup}

\paragraph{Downstream Dataset Construction.} For the downstream prediction experiments, we construct a machine learning dataset from the building-level inventory file. We restrict the dataset to buildings with available Scope 3 emission targets, and we require each building to have a valid satellite image path and successfully parsed metadata. 

\paragraph{Input Features.} The inputs consist of tabular building metadata and satellite imagery. Tabular features cover building geometry, geographic context, type, energy activity, temporal indicators, and provenance flags. We use two feature sets in our experiments:
\begin{itemize}[noitemsep, topsep=0pt]
    \item \textbf{Basic features}: floor area + geographic coordinates only, used as a minimal-feature ablation.
    \item \textbf{Full features}: the complete Scope3Trace tabular schema enumerated in Appendix~\ref{subsec:full_features}.
\end{itemize}

These attributes capture basic geometric and contextual properties of buildings.

\paragraph{Satellite Imagery Pipeline.} Each building's standardized address is geocoded via the Google Maps Static API and the resulting coordinates are passed to the \textbf{Mapbox API} to retrieve satellite imagery at \textbf{1024$\times$1024 resolution (2$\times$ retina)}. Images are encoded using a frozen ImageNet-pretrained \textbf{ResNet-50} that produces \textbf{2048-dimensional embeddings}. These embeddings are concatenated with the tabular features and fed into the downstream regression model.

\paragraph{Prediction Targets.} We evaluate two prediction tasks, the first task is to predict the total Scope 3 emissions for each building. Due to the highly skewed distribution of emissions, the target variable was log-transformed during training. During evaluation, the predicted results will be converted back to their original scale. The second task is to predict Scope 3 emissions at the category level. This task is formulated as a multi-output regression problem where the model predicts all categories simultaneously. For category prediction, we have also included the logarithmically transformed total emissions as an auxiliary feature in the tabular input.

\subsubsection{Results}

Interpretation of absolute performance should account for target noise:
96.7\% of building-level Scope~3 targets are proxy-derived
(Table~\ref{tab:provenance}), and commercial emissions-data vendors
agree with one another on Scope~3 at correlations as low as 0.22, that
is, $R^2 \approx 0.05$ \citep{busch2022}. Reported $R^2$ values should
therefore be read against this consistency ceiling rather than as
absolute predictive accuracy. We evaluated the predictive capability of this dataset for two downstream tasks. Detailed numerical result can be found in Appendix~\ref{app:prediction_results}. 

Overall, the results indicate that the dataset contains meaningful predictive signals applicable to both tasks. The models were able to learn patterns that support the estimation of total Scope 3 emissions and the prediction of individual emission categories, suggesting that the constructed dataset provides sufficient information for building-level emissions modeling.

For both tasks, models utilizing satellite imagery consistently reduced prediction error in the log-transformed target space compared to baseline models using tabular data alone. This suggests that visual information captures contextual signals related to the built environment that are not fully reflected in structured building metadata.

Furthermore, multimodal models combining tabular attributes with image representations demonstrated the strongest overall performance. This indicates that building metadata and satellite imagery provide complementary information for emissions estimation.

Taken together, these results demonstrate that the proposed dataset supports both the prediction of total Scope 3 emissions and the prediction of emissions across various categories, and provides a useful benchmark for multimodal emissions modeling.

\section{Prediction Benchmark Results}
\label{app:prediction_results}

Tables~\ref{tab:total_prediction_results} and~\ref{tab:category_prediction_results} report results for total and category-level Scope~3 prediction respectively. We report two settings:
\begin{itemize}[noitemsep, topsep=0pt]
    \item \textbf{Basic features}: floor area + geographic coordinates only, used to isolate the contribution of modalities.
    \item \textbf{Full features}: all Scope3Trace attributes (basic features plus building usage category, footprint area, number of levels, energy data, and provenance metadata).
\end{itemize}

Under the full Scope3Trace feature set, the multimodal model achieves
$R^2 = \textbf{0.46}$ for total Scope~3 prediction and macro
$R^2 = \textbf{0.54}$ for category-level prediction. While modest in
absolute terms, these values should be read against the reliability of
currently available Scope~3 data, where commercial data vendors agree
with one another at correlations as low as 0.22 ($R^2 \approx 0.05$)
\citep{busch2022}; relative to the basic-feature
baselines, the dataset thus provides useful predictive signal for
downstream emissions modeling, rather than strong absolute predictive
performance.

\begin{table}[t]
\centering
\small
\begin{tabular}{lcc}
\toprule
Model & log-RMSE & $R^2$ \\
\midrule
\multicolumn{3}{l}{\emph{Basic features (floor area + coords)}} \\
Tabular-only & 1.38 $\pm$ 0.02 & 0.03 $\pm$ 0.01 \\
Image-only & 1.24 $\pm$ 0.03 & 0.01 $\pm$ 0.01 \\
Tabular + Image & 1.24 $\pm$ 0.01 & 0.04 $\pm$ 0.01 \\
\midrule
\multicolumn{3}{l}{\emph{Full Scope3Trace features}} \\
MLP (tabular only) & 1.05 $\pm$ 0.02 & 0.42 $\pm$ 0.02 \\
MLP (tabular + image) & \textbf{1.02 $\pm$ 0.02} & \textbf{0.46 $\pm$ 0.02} \\
\bottomrule
\end{tabular}
\caption{Prediction performance for total Scope~3 emissions under basic-feature ablations and the full Scope3Trace feature set. Mean $\pm$ standard deviation over 3 runs.}
\label{tab:total_prediction_results}
\end{table}

\begin{table}[t]
\centering
\small
\begin{tabular}{lcc}
\toprule
Model & log-RMSE & macro $R^2$ \\
\midrule
\multicolumn{3}{l}{\emph{Basic features (floor area + coords)}} \\
Ridge & 1.97 $\pm$ 0.03 & 0.00 $\pm$ 0.01 \\
Random Forest & 2.06 $\pm$ 0.04 & -0.76 $\pm$ 0.05 \\
XGBoost & 2.19 $\pm$ 0.05 & -2.21 $\pm$ 0.08 \\
Tabular MLP & 1.72 $\pm$ 0.02 & 0.08 $\pm$ 0.02 \\
Image-only & 1.96 $\pm$ 0.03 & -0.01 $\pm$ 0.02 \\
Tabular + Image & 1.72 $\pm$ 0.01 & 0.11 $\pm$ 0.02 \\
\midrule
\multicolumn{3}{l}{\emph{Full Scope3Trace features}} \\
MLP (tabular only) & 1.20 $\pm$ 0.02 & 0.50 $\pm$ 0.02 \\
MLP (tabular + image) & \textbf{1.15 $\pm$ 0.02} & \textbf{0.54 $\pm$ 0.02} \\
\bottomrule
\end{tabular}
\caption{Macro-averaged prediction performance for category-level Scope~3 emissions under basic-feature ablations and the full Scope3Trace feature set. Mean $\pm$ standard deviation over 3 runs.}
\label{tab:category_prediction_results}
\end{table}

\subsection{Full Input Feature Specification}
\label{subsec:full_features}

Table~\ref{tab:full_features} lists every tabular field used in the
\emph{full features} setting of our prediction benchmark. Fields are
grouped by category and annotated with their source within the
Scope3Trace schema (cf.\ Appendix~\ref{appendix:dataset_schema}).
For categorical fields (e.g., building usage), one-hot encoding is
applied before model input; numeric fields are standardized to
zero mean and unit variance per training fold.

\begin{table}[t]
\centering
\footnotesize
\setlength{\tabcolsep}{6pt}
\caption{Full tabular input features used in the \emph{full features}
setting of the building-level Scope~3 prediction benchmark
(Tables~\ref{tab:total_prediction_results}
and~\ref{tab:category_prediction_results}).
Numeric fields are standardized per training fold;
categorical fields are one-hot encoded;
binary fields are mapped to $\{0,1\}$.
``log1p'' denotes $\log(1+x)$ applied before standardization.}
\label{tab:full_features}
\begin{tabularx}{\columnwidth}{>{\RaggedRight\arraybackslash}X l}
\toprule
Field (source column) & Preprocessing \\
\midrule
\multicolumn{2}{l}{\textit{(a) Basic building attributes in the full feature set}} \\
\midrule
\texttt{floor\_area} (\texttt{gba\_floor\_area\_final\_m2}, fallback \texttt{floor\_area\_sqm}) & log1p, num. \\
\texttt{footprint\_area} (\texttt{gba\_footprint\_area\_m2}) & log1p, num. \\
\texttt{levels} (\texttt{gba\_levels\_used}) & log1p, num. \\
\texttt{lat}, \texttt{lon} (\texttt{geo\_point\_2d} / \texttt{location\_lat\_lon}) & Numeric \\
\texttt{usage} $\rightarrow$ \{commercial, residential\} & One-hot (2-d) \\
\texttt{year\_built} & Numeric \\
\midrule
\multicolumn{2}{l}{\textit{(b) NABERS-derived features}} \\
\midrule
\texttt{nabers\_ghg\_with\_re\_kgco2\_m2} (intensity, incl.\ renewables) & Numeric \\
\texttt{nabers\_ghg\_without\_re\_kgco2\_m2} (intensity, excl.\ renewables) & Numeric \\
\texttt{nabers\_greenpower\_ratio} (renewable share) & Numeric \\
\texttt{nabers\_total\_renewable\_kwh} & Numeric \\
\texttt{nabers\_nonrenewable\_kwh} & Numeric \\
\texttt{nabers\_star\_rating} & Numeric \\
\texttt{nabers\_onsite\_solar\_kw} (rooftop PV capacity) & Numeric \\
\texttt{nabers\_carbon\_neutral} (YES/NO) & Binary $\{0,1\}$ \\
\texttt{nabers\_building\_type} (Office, Shopping Centre, Hotel, \ldots) & Categorical \\
\midrule
\multicolumn{2}{l}{\textit{(c) Scope~1/2 emissions (excluded in ablation)}} \\ \\
\midrule
\texttt{scope1\_tco2e} & Numeric \\
\texttt{scope2\_location\_tco2e} & Numeric \\
\texttt{scope2\_market\_tco2e} & Numeric \\
\midrule
\multicolumn{2}{l}{\textit{(d) Energy use breakdown}} \\
\midrule
\texttt{energy\_electricity\_kwh} & Numeric \\
\texttt{energy\_gas\_mj} & Numeric \\
\texttt{energy\_renewables\_kwh} & Numeric \\
\texttt{energy\_residual\_kwh} & Numeric \\
\texttt{energy\_total\_kwh} & Numeric \\
\texttt{tenant\_electricity\_kwh} & Numeric \\
\texttt{p\_com}, \texttt{p\_res} (commercial / residential share) & Numeric \\
\midrule
\multicolumn{2}{l}{\textit{(e) Multimodal feature (concatenated with tabular features)}} \\
\midrule
ResNet-50 embedding (frozen, ImageNet-pretrained, 1024$\times$1024 at 2$\times$ retina) & 2048-d vector \\
\bottomrule
\end{tabularx}
\end{table}

\section{Artifact Licenses and Intended Use}
\label{app:licenses}

\paragraph{Released.}
The pipeline code and baseline configurations are released under the MIT
license. The gold evaluation set with page-level evidence anchors and the
emission-factor lookup tables are released publicly under CC BY 4.0 to
support third-party auditing. The full Scope3Trace dataset is archived on
Zenodo under restricted access and available to researchers on request
for research use. Rights to the verbatim evidence excerpts remain with
the original publishers; their sole function is verification via the
released source URLs. Rights holders may request removal of specific
excerpts by contacting the corresponding author; affected records retain
their values and page anchors with the excerpt text redacted. The
dataset is not intended to replace audited disclosures or support
compliance reporting.

\paragraph{Not redistributed.} Raw source PDFs (source URLs and short evidence snippets only), raw Mapbox/Google Maps imagery (only frozen ResNet-50 embeddings), and raw LLM API outputs.

\paragraph{Inputs.} All external artifacts are used within the provider-specified intended scope, and no raw provider data is redistributed. Specifically: ESG / sustainability PDFs are public corporate disclosures, from which we retain source URLs and short page-anchored evidence snippets only; NABERS, NGER, and Climate Active are public Australian government data, used as derived building-level features; Google Maps Geocoding (Maps Platform ToS) is used to obtain (lat, lon) coordinates only; Mapbox Static Images (Mapbox ToS) is used to retrieve satellite imagery for offline feature extraction, and only frozen ResNet-50 embeddings are released; ResNet-50 (\texttt{torchvision}, ImageNet-pretrained) is used as a frozen feature extractor; and LLM APIs (GPT-4o, GPT-4o-mini, Gemini-2.0, DeepSeek-V3) are called under their respective vendor terms of service at temperature=0, with no raw API outputs released.

\end{document}